\documentclass[journal,twoside,web]{ieeecolor}
\usepackage{tmi}
\usepackage{cite}
\usepackage[caption=false]{subfig}
\usepackage{amsmath,amssymb,amsfonts}
\usepackage{algorithmic}
\usepackage[flushleft]{threeparttable}
\usepackage{float}
\usepackage{multirow,makecell,multicol}
\usepackage{graphicx}
\usepackage{textcomp}
\usepackage{pifont}
\usepackage[pagebackref=false,breaklinks=true,letterpaper=true,colorlinks=false,bookmarks=false]{hyperref}
\def\figurename{Fig.}
\def\sectionname{Sec.}
\def\tablename{Table}

\newcommand{\etal}{\mbox{\emph{et al.\ }}}
\newcommand{\ie}{\mbox{\emph{i.e.,\ }}}
\newcommand{\eg}{\mbox{\emph{e.g.,\ }}}

\newcommand{\cmark}{\ding{51}}%
\newcommand{\xmark}{\ding{53}}%
\usepackage{array}
\usepackage[table]{xcolor}
\usepackage{xcolor}
\usepackage[normalem]{ulem} 

\definecolor{maroon}{RGB}{58, 183, 149}
\definecolor{iblue}{RGB}{33, 53, 236 } 

\definecolor{mred}{RGB}{255, 0, 0 }
\newcolumntype{P}[1]{>{\centering\arraybackslash}p{#1}}

\def\BibTeX{{\rm B\kern-.05em{\sc i\kern-.025em b}\kern-.08em
    T\kern-.1667em\lower.7ex\hbox{E}\kern-.125emX}}
\begin{document}

\title{Transferable Visual Words: \\ Exploiting the Semantics of Anatomical Patterns for Self-supervised Learning}

\author{Fatemeh Haghighi,  Mohammad Reza Hosseinzadeh Taher, Zongwei Zhou, \IEEEmembership{Member, IEEE}, Michael B. Gotway, and Jianming Liang, \IEEEmembership{Senior Member, IEEE},

\thanks{F. Haghighi and M.R. Hosseinzadeh Taher are with  School of Computing, Informatics, and Decision Systems Engineering, Arizona State University, Tempe, AZ 85281 USA (e-mail:fhaghigh@asu.edu, mhossei2@asu.edu). }
\thanks{Z. Zhou and J. Liang are with  Department of Biomedical
Informatics, Arizona State University, Scottsdale, AZ 85259 USA (e-mail: zongweiz@asu.edu,jianming.liang@asu.edu).}
\thanks{M. B. Gotway is with the Radiology Department,
Mayo Clinic, Scottsdale, AZ 85259 USA (e-mail: Gotway.Michael@mayo.edu).}}

\maketitle

\begin{abstract}
This paper introduces a new concept called ``transferable visual words'' (TransVW), aiming to achieve annotation efficiency for deep learning in medical image analysis. Medical imaging---focusing on particular parts of the body for defined clinical purposes---generates images of great similarity in anatomy across patients and yields sophisticated anatomical patterns across images, which are associated with rich {\em semantics} about human anatomy and which are natural {\em visual words}. We show that these visual words can be automatically harvested according to anatomical consistency via self-discovery, and that the self-discovered visual words can serve as strong yet free supervision signals for deep models to learn semantics-enriched generic image representation via self-supervision (self-classification and self-restoration). 
Our extensive experiments demonstrate the annotation efficiency of TransVW by offering higher performance and faster convergence with reduced annotation cost in several applications. 
Our TransVW has several important advantages, including (1) TransVW is a fully autodidactic scheme, which exploits the semantics of visual words for self-supervised learning, requiring no expert annotation; (2) visual word learning is an add-on strategy, which complements existing self-supervised methods, boosting their performance; and (3) the learned image representation is semantics-enriched models, which have proven to be more robust and generalizable, saving annotation efforts for a variety of applications through transfer learning.
Our code, pre-trained models, and curated visual words are available at \href{https://github.com/JLiangLab/TransVW}{https://github.com/JLiangLab/TransVW}.
\end{abstract}

\begin{IEEEkeywords}
Self-supervised learning, transfer learning, visual words, anatomical patterns, computational anatomy, 3D medical imaging, and 3D pre-trained models.
\end{IEEEkeywords}

\section{Introduction}
\label{sec:introduction}
\IEEEPARstart{A}{} grand promise of computer vision is to learn general-purpose image representation, seeking to automatically discover generalizable knowledge from data in either a supervised or unsupervised manner and transferring the discovered knowledge to a variety of applications for performance improvement and annotation efficiency. 

In the literature, convolutional neural networks (CNNs) and bags of visual words (BoVW)~\cite{Sivic2003Video} are often presumed to be competing methods, but they actually offer complementary strengths. Training CNNs requires a large number of annotated images, but the learned features are transferable to many applications. Extracting visual words in BoVW, on the other hand, is unsupervised in nature, demanding no expert annotation, but the extracted visual words lack transferability. Therefore, the first question that we seek to answer is \textit{how to beneficially integrate the transfer learning capability of CNNs with the unsupervised nature of BoVW in extracting visual words for image representation learning?}

\begin{figure}[t]
\centerline{\includegraphics[width=0.95\columnwidth]{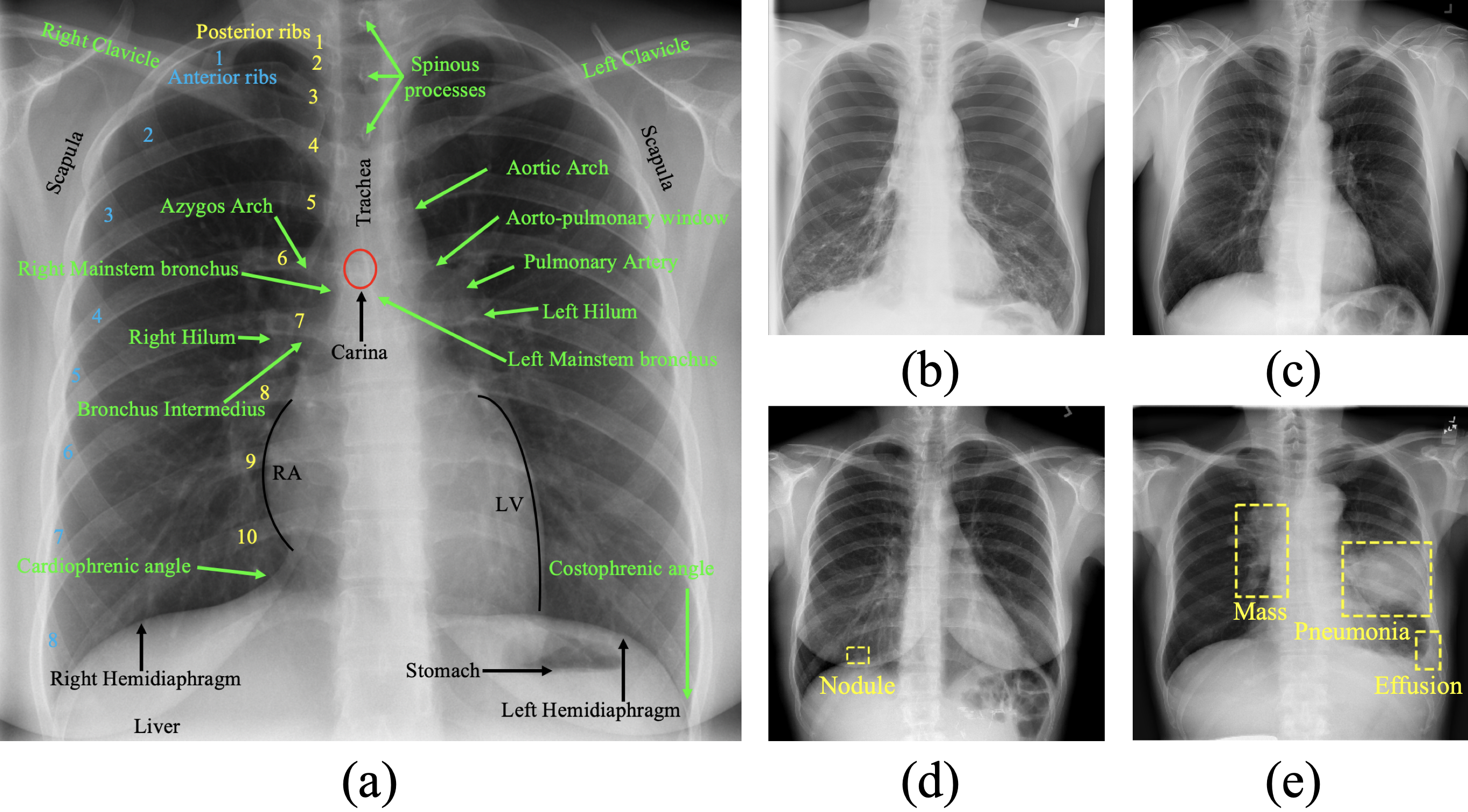}}
\caption{ Without loss of generality, we illustrate our idea in 2D with chest X-rays. The great similarity of the lungs in anatomy, partially annotated in (a), across patients yields complex yet consistent and recurring anatomical patterns across X-rays in healthy (a, b, and c) or diseased (d and e), which we  refer to as anatomical \textit{visual words}. Our proposed TransVW (transferable visual words) aims to learn {\em generalizable} image representation from the anatomical visual words {\em without} expert annotations, and {\em transfer} the learned deep models to create powerful application-specific target models. TransVW is general and applicable across organs, diseases, and modalities in both 2D and 3D.}
\label{fig:annotated_xray}
\end{figure}

\begin{figure*}[t]
\centerline{\includegraphics[width=0.95\textwidth]{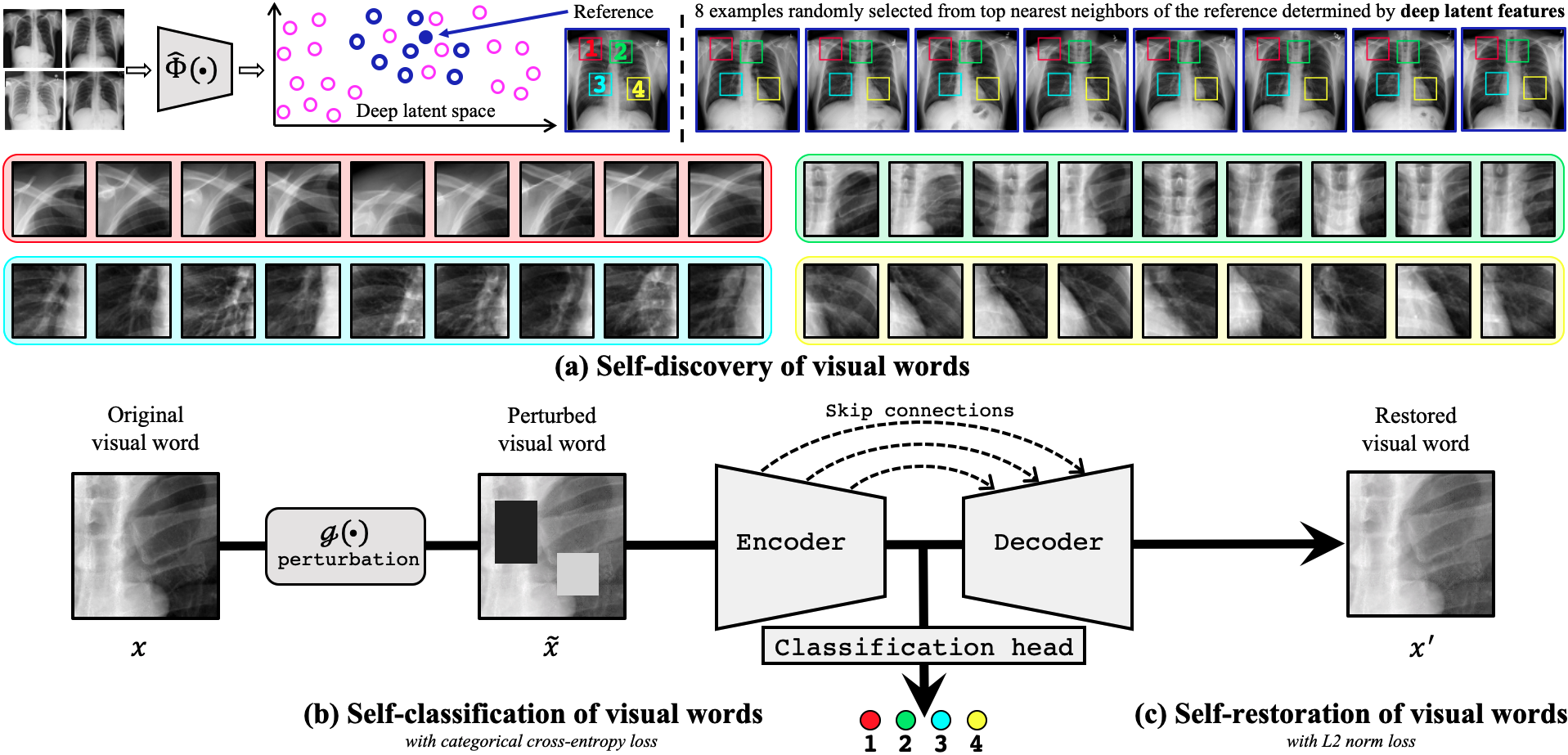}}
\caption{Our proposed self-supervised learning framework TransVW is for learning general-purpose image representation enriched with the semantics of anatomical visual words by (a) self-discovery, (b) self-classification, and (c) self-restoration. First, to discover anatomically consistent instances for each visual word across patients, we train a feature extractor $\hat{\Phi}(.)$ (\eg auto-encoder) with unlabeled images, so that images of great resemblance can be automatically identified based on its deep latent features. Second, after selecting a random reference patient and using the feature extractor to find  patients similar in appearance, to extract instances of a visual word, we crop image patches at a random yet fixed coordinate across all selected patients and assign a unique (pseudo) label to the extracted patches (instances). For simplicity and clarity, we have shown instances of four visual words extracted at four different random coordinates to illustrate the similarity and consistency among the discovered instances of each visual word. Our self-discovery automatically curates a set of visual words associated with semantically meaningful labels, providing a free and rich source for training deep models to learn semantic representations. 
Finally, we perturb instances of the visual words with $\mathrm{g}(.)$ and give them as input to an encoder-decoder network with 
skip connections in between and a classification head at the end of the encoder. Our self-classification and self-restoration of visual words empower the deep model to learn anatomical semantics from the visual words, resulting in image representation, which has proven to be more generalizable and transferable to a variety of target tasks. 
}
\label{fig:method}
\end{figure*}

In the meantime, medical imaging protocols, typically designed for specific clinical purposes by focusing on particular parts of the body, generate images of great similarity in anatomy across patients and yield an abundance of sophisticated anatomical patterns across images. These anatomical patterns are naturally associated with rich semantic knowledge about human anatomy. Therefore, the second question that we seek to answer is \textit{how to exploit the deep semantics associated with anatomical patterns embedded in medical images to enrich image representation learning?}

In addressing the two questions simultaneously, we have conceived a new concept: \textbf{transferable visual words (TransVW)}, where the sophisticated anatomical patterns across medical images are natural ``visual words'' associated with deep semantics in human anatomy (see \figurename~\ref{fig:annotated_xray}). These anatomical visual words can be automatically harvested from unlabeled medical images and serve as strong yet free supervision signals for CNNs to learn semantics-enriched representation via self-supervision. The learned representation is generalizable and transferable because it is not biased to the idiosyncrasies of pre-training (pretext) tasks and datasets, thereby it can produce more powerful models to solve application-specific  (target) tasks via transfer learning. As illustrated in \figurename~\ref{fig:method}, our TransVW consists of three components: 
(1) a novel self-discovery scheme that automatically harvests anatomical visual words, exhibiting consistency (in pattern appearances  and  semantics) and diversity (in organ shapes, boundaries, and textures), directly from unlabeled medical images and assigns each of them with a unique label that bears the semantics associated with a particular part of the body;  (2) a unique self-classification scheme that compels the model to learn semantics from anatomical consistency within visual words; and (3) a scalable self-restoration scheme that encourages the model to encode anatomical diversity within visual words.

Our extensive experiments demonstrate the annotation efficiency of TransVW in higher performance, faster convergence, and less annotation cost on the applications where there is a dearth of annotated images. 
Compared with existing publicly available models,  pre-trained by either self-supervision or full-supervision, our TransVW offers several advantages, including (1) TransVW is a fully autodidactic scheme, which exploits the semantics of visual words for self-supervised learning, requiring no expert annotation; (2) visual word learning is an add-on strategy, which complements existing self-supervised methods, boosting their performance; and (3) the learned image representation is semantics-enriched models, which have proven to be more robust and generalizable, saving annotation efforts for a variety of applications through transfer learning. In summary, we make the following contributions:
\begin{itemize}
    \item An unsupervised clustering strategy, curating a dataset of anatomical visual words from unlabeled medical images.
    \item An add-on learning scheme, enriching representations learned from existing self-supervised methods~\cite{gidaris2018unsupervised,pathak2016context,chen2019self,zhou2019models}.
    \item An advanced self-supervised framework, elevating transfer learning performance, accelerating training speed, and reducing annotation efforts.
\end{itemize}

\section{Transferable Visual Words}
\label{sec:method}
TransVW aims to learn transferable and generalizable image representation by leveraging the semantics associated with the anatomical patterns embedded in medical images (see \figurename~\ref{fig:annotated_xray}). For clarity, as illustrated in \figurename~\ref{fig:method}a, we define a \textit{visual word} as a segment of consistent anatomy recurring across images and the \textit{instances of a visual word} as the patches extracted across different but resembling images for this visual word. Naturally, all instances of a visual word exhibit great similarity in both appearance and semantics. Furthermore, to reflect the semantics of its corresponding anatomical parts, a unique (pseudo) label is automatically assigned to each visual word during the self-discovery process (see Section~\ref{sec:vw_discovery}); consequently, all instances of a visual word share the same label bearing the same semantics in anatomy.

\subsection{TransVW learns semantics-enriched representation}
\label{sec:method_semantics-enriched}

TransVW enriches representation learning with the semantics of visual words, as shown in \figurename~\ref{fig:method}, through the following three components.

\smallskip
\noindent\textbf{1) Self-discovery---harvesting the semantics of anatomical patterns to form visual words.}
\label{sec:vw_discovery}
The self-discovery component aims to automatically extract a set of $C$ visual words from unlabeled medical images as shown in~\figurename~\ref{fig:method}a. To ensure a high degree of consistency in anatomy among the instances of each visual word, we first identify a set of $K$ patients that display a great similarity in their overall image appearance. To do so, we use the whole patient scans in the training dataset to train a feature extractor $\hat{\Phi}(.)$, an auto-encoder network, to learn an identical mapping from a whole-patient scan to itself. Once trained, its latent features can be used as an indicator of the similarity among patient scans. As a result, we can form such a set of $K$ patient scans by randomly anchoring a patient scan as the reference and appending its top $K$--1 nearest neighbors found throughout the entire training dataset based on the $L2$ distance in the latent feature space. Given the great resemblance among the selected $K$ patient scans, the patches extracted at the same coordinate across these $K$ scans are expected to exhibit a high degree of similarity in anatomical patterns. Therefore, for each visual word, we extract its $K$ instances by cropping around a random but fixed coordinate across a set of selected $K$ patients, and assign a unique pseudo label to them. We repeat this process $C$ times, yielding a set of $C$ visual words, each with $K$ instances, extracted from $C$ random coordinates. The extracted visual words are naturally associated with the semantics of the corresponding human body parts. For example, in~\figurename~\ref{fig:method}a, four visual words are defined randomly in a reference patient  (top-left), bearing local anatomical information of (1) anterior ribs 2--4, (2) spinous processes, (3) right pulmonary artery, and (4) LV.
In summary, our self-discovery automatically generates a dataset of visual words associated with semantic labels, as a free and rich source for training deep models to learn semantics-enriched representations from unlabeled medical images

\medskip
\noindent\textbf{2) Self-classification---learning the semantics of anatomical consistency from visual words.} 
\label{sec:vw_classification}
Once a set of visual words are self-discovered, representation learning can be formulated as self-classification, a $C$-way multi-class classification that discriminates visual words based on their semantic (pseudo) labels.  
As illustrated in~\figurename~\ref{fig:method}b, the self-classification branch is composed of (1) an encoder that projects visual words into a latent space, and (2) a classification head, a sequence of fully-connected layers, that predicts the pseudo label of visual words. 
It is trained by minimizing the standard categorical cross-entropy loss function: 
\begin{equation}\label{eq:cls_loss}
\mathcal{L}_{cls}=-\frac{1}{B}\sum_{b=1}^{B}\sum_{c=1}^{C}\mathcal{Y}_{bc}\log  {\mathcal{P}_{bc}}
\end{equation}
where $B$ denotes the batch size; $C$ denotes the number of visual words; $\mathcal{Y}$ and $\mathcal{P}$ represent the ground truth (one-hot pseudo label vector) and the network prediction, respectively. Through training, the model is compelled to learn features that distinguish the anatomical dissimilarity among instances belonging to different visual words and that recognize the anatomical resemblance among instances belonging to the same visual words, resulting in image representations associated with the semantics of anatomical patterns underneath medical images. Therefore, self-classification is for learning image representation enriched with semantics that can pull together all instances of each visual word, while pushing apart instances of different visual words.

\medskip
\noindent\textbf{3) Self-restoration---encoding the semantics of anatomical diversity of visual words.}
\label{sec:vw_restoration}
In self-discovery,  we intentionally selected patients, rather than patches, according to their resemblance at the whole patient level. Given that no scans of different patients are the same in appearance, behind their great similarity, the instances of a visual word are also expected to display subtle anatomical diversity in terms of organ shapes, boundaries, and texture. Such a balance between the consistency and diversity of anatomical patterns for each visual word is critical for deep models to learn robust image representation. To encode this anatomical diversity within visual words for image representation learning, we augment our framework with self-restoration, training the model to restore the original visual words from the perturbed ones. 
The self-restoration branch, as shown in~\figurename~\ref{fig:method}c, is an encoder-decoder with skip connections in between. 
We apply a perturbation operator $\mathrm{g}(.)$ on a visual word $x$ to get the perturbed visual word $\tilde{x}=\mathrm{g}(x)$.  The encoder takes the perturbed visual word $\tilde{x}$  as an input and generates a latent representation. The decoder takes the latent representation from the encoder and decodes it to produce the original visual word. 
Our perturbation operator $\mathrm{g}(.)$ consists of non-linear, local-shuffling, out-painting, and inpainting transformations, which are proposed by Zhou \etal~\cite{zhou2019models,zhou2021models}, as well as identity mapping (\ie $x=\mathrm{g}(x)$).  Restoring visual word instances from their perturbations enables the model to learn image representation from multiple perspectives~\cite{zhou2019models,zhou2021models}.  
The restoration branch is trained by minimizing the $L2$ distance between the original and reconstructed visual words: 

\begin{equation}\label{eq:cls_loss}
\mathcal{L}_{rec}=\frac{1}{B}\sum_{i=1}^{B}\|x_i - x_i'\|_2
\end{equation}
where $B$ denotes the batch size, $x$ and $x'$ represent the  original visual word and the reconstructed prediction, respectively.  

\bigskip
To enable end-to-end representation learning from multiple sources of information and yield more powerful models for a variety of medical tasks, in TransVW, self-classification and self-restoration are integrated together by sharing the encoder and jointly trained with one single objective function: 
\begin{equation}
\mathcal{L}=\lambda_{cls}\mathcal{L}_{cls}+\lambda_{rec}\mathcal{L}_{rec} 
\end{equation}
where $\lambda_{cls}$ and $\lambda_{rec}$ adjust the weights of classification and restoration losses, respectively.
Our {\em unique} definition of $\mathcal{L}_{cls}$  empowers the model  to learn the common anatomical semantics across medical images from a strong discriminative signal---the semantic label of visual words.
The definition of $\mathcal{L}_{rec}$ equips the model to learn the anatomical finer details of visual words from multiple perspectives by restoring original visual words from varying image perturbations.
We should emphasize that our goal is {\em not} simply discovering, classifying, and restoring visual words \textit{per se}, rather advocating it as a holistic pre-training scheme for learning semantics-enriched image representation, whose usefulness must be assessed objectively based on its generalizability and transferability to various target tasks.

\begin{table*}[t]
\begin{center}
\begin{threeparttable}
\caption{
Target tasks for Transfer Learning.  We transfer the learned representations by fine-tuning it for seven medical imaging applications including 3D and 2D image classification and segmentation tasks. To evaluate the generalization ability of our TransVW,  we select a diverse set of applications ranging from  the  tasks  on  the  same  dataset  as  pre-training  to  the tasks  on  the  unseen  organs,  datasets,  or  modalities  during pre-training. For each task, \cmark denotes the properties that are in common between the pretext and target tasks.  
}
\label{tab:datasets}
\begin{tabular}{p{0.03\linewidth}p{0.29\linewidth}p{0.12\linewidth} p{0.05\linewidth}p{0.14\linewidth} | P{0.05\linewidth} P{0.05\linewidth} P{0.05\linewidth}  }
\hline
\multirow{2}{*}{Code$^{\ast}$}  & \multirow{2}{*}{Application}  & \multirow{2}{*}{Object}  & \multirow{2}{*}{Modality}  & \multirow{2}{*}{Dataset}   & \multicolumn{3}{c}{Common with pretext task}\\
\cline{6-8} & & & & & Organ & Dataset& Modality\\
\hline
\texttt{NCC} & Lung nodule false positive reduction &Lung Nodule &CT & LUNA16~\cite{setio2017validation}& \cmark & \cmark & \cmark\\
\texttt{NCS} & Lung nodule segmentation  &Lung Nodule  &CT & LIDC-IDRI~\cite{armato2011lung} & \cmark & \cmark & \cmark\\
\texttt{ECC} & Pulmonary embolism false positive reduction & Pulmonary Emboli & CT & PE-CAD~\cite{tajbakhsh2015computer}& \cmark &  & \cmark\\
\texttt{LCS}& Liver segmentation  &Liver &CT & LiTS-2017~\cite{bilic2019liver} & & & \cmark\\
\texttt{BMS}& Brain Tumor Segmentation  &Brain Tumor &MRI & BraTS2018~\cite{bakas2018identifying} \\
\hline
\texttt{DXC}& Fourteen thorax diseases classification  & Thorax Diseases&X-ray & ChestX-Ray14~\cite{wang2017chestx} & \cmark & \cmark & \cmark\\
\texttt{\texttt{PXS}}& Pneumothorax Segmentation &Pneumothorax &X-ray& SIIM-ACR-2019~\cite{PNEchallenge}& \cmark &  & \cmark \\
\hline
\end{tabular}

\begin{tablenotes}
        \item $^{\ast}$ The first letter denotes the object of interest (``\texttt{N}'' for lung nodule, ``\texttt{B}'' for brain tumor, ``\texttt{L}'' for liver, etc); the second letter denotes the modality (``\texttt{C}'' for CT, ``\texttt{X}'' for X-ray, ``\texttt{M}'' for MRI);  the last letter denotes the task (``\texttt{C}'' for classification, ``\texttt{S}'' for segmentation).
\end{tablenotes}
\end{threeparttable}
\end{center}
\end{table*}

\subsection{TransVW has several unique properties}

\noindent\textbf{1) Autodidactic---exploiting semantics in unlabeled data for self supervision.} 
Due to the lack of sufficiently large, curated, and labeled medical datasets, self-supervised learning holds a great promise for representation learning in medical imaging because it does not require manual annotation for pre-training. Unlike existing self-supervised methods for medical imaging~\cite{chen2019self,Zhuang2019Self,zhou2019models}, our TransVW explicitly employs the (pseudo) labels that bear the semantics associated with the sophisticated anatomical patterns embedded in the unlabeled images to learn more pronounced representation for medical applications. Particularly, TransVW benefits from  a large, diverse set of anatomical visual words discovered by our self-discovery process, coupled with a training scheme integrating both self-classification and self-restoration, to learn semantics-enriched representation from unlabeled medical images. With \textit{zero} annotation cost, our TransVW not only outperforms other self-supervised methods but also surpasses publicly-available, fully-supervised pre-trained models, such as I3D~\cite{carreira2017quo}, NiftyNet~\cite{gibson2018niftynet}, and MedicalNet~\cite{chen2019med3d}.

\medskip
\noindent\textbf{2) Comprehensive---blending consistency with diversity for semantics richness.}
Our self-discovery component secures both consistency and diversity within each visual word, thereby offering a lucrative source for pre-training deep models. More specifically, our self-discovery process computes similarity at the patient level and selects the top nearest neighbors of the reference patient. Extracting visual word instances from these similar patients, based on random but fixed coordinates, strikes a balance between consistency and diversity of the anatomical pattern within each visual word. Consequently, our self-classification exploits the semantic consistency by classifying visual words according to their pseudo labels, resulting in class-level feature separation among visual word classes. Furthermore, our self-restoration leverages the fine-grained anatomical information—--the subtle diversity of intensity, shape, boundary, and texture enables instance-level feature separation among instances within each visual word. As a result, our TransVW projects visual words into more comprehensive feature space in both class-level and instance-level by blending consistency with diversity.

\medskip
\noindent\textbf{3) Robust---preventing superficial solutions for deep representation.} 
Self-supervised learning is notorious for learning shortcut solutions in tackling pretext tasks, leading to less generalizable image representations~\cite{Jenni2020Steering,doersch2015unsupervised}. However, our method, especially our self-classification component, is discouraged from learning superficial solutions since our self-discovery process imposed substantial diversity among instances of each visual word. Furthermore, we follow two well-known techniques to further improve the diversity of visual words. Firstly, following Doersch \etal~\cite{doersch2015unsupervised} and Mundhenk \etal~\cite{mundhenk2018improvements}, we extract multi-scale instances for each visual word within a patient, in which each instance is randomly jittered by a few pixels. Consequently, having various scale instances in each class enforces self-classification to perform more semantic reasoning by preventing easy matching of simple features among the instances of the same class~\cite{mundhenk2018improvements}. Secondly, during pre-training, we augment visual words with various image perturbations to increase the diversity of data. 
Altogether,  the  substantial  diversity  of  visual  words  coupled with various image perturbations enforce our pretext task to capture semantic-bearing features, resulting in a compelling and robust representation obtained from anatomical visual words.

\medskip
\noindent\textbf{4) Versatile---complementing existing self-supervised methods for performance enhancement.} 
TransVW boasts an innovative add-on capability, a versatile feature that is unavailable in other self-supervised learning methods. Unlike existing self-supervised methods that build supervision merely from the information within individual images of training data, our self-discovery and self-classification leverage the anatomical similarities present across different images (reflected in visual words) to learn common anatomical semantics. Consequently, incorporating visual word learning into existing self-supervised methods enforces them to encode semantic structures of visual words into their learned embedding space, resulting in more versatile representations. Therefore, our self-discovery and self-classification components together can serve as an add-on to boost existing self-supervised methods, as evidenced in \figurename~\ref{fig:w_wo_semantics}.

\section{Experiments}
\subsection{Pre-training TransVW}
\label{sec:Pre-training_TransVW}
TransVW models are pre-trained solely on unlabeled images. Nevertheless, to ensure no test-case leaks from pretext tasks to target tasks, any images that will be used for validation or test in target tasks (see \tablename~\ref{tab:datasets}) are excluded from pre-training. 
We have released two pre-trained models: (1) TransVW Chest CT in 3D, which was pre-trained from scratch using 623 chest CT scans in LUNA16~\cite{setio2017validation} (the same as the publicly released Models Genesis\footnote{\label{foot:modelsgenesis}Models Genesis: \href{https://github.com/MrGiovanni/ModelsGenesis}{github.com/MrGiovanni/ModelsGenesis}}), and (2) TransVW Chest X-rays in 2D, which was pre-trained on 76K chest X-ray images in ChestX-ray14~\cite{wang2017chestx} datasets. We set  $C=45$ for TransVW Chest CT and $C=100$  for TransVW Chest X-rays (see Appendix~\ref{appendix_abl_num_vws} for the ablation studies on the impact of the number of visual words on performance).   
We empirically set $K=200$  and $K=1000$ for TransVW Chest CT and TransVW Chest X-rays, respectively, to strike a balance between diversity and consistency of the visual words. For each instance of a visual word, multi-scale cubes/patches for 3D/2D images are cropped, and then all are resized to 64$\times$64$\times$32 and 224$\times$224 for TransVW Chest CT and TransVW Chest X-rays, respectively (see Appendix~\ref{appendix_vws} for samples of the discovered visual words).

\subsection{Fine-tuning TransVW}
\label{sec:experiments_finetune}

The pre-trained TransVW can be used for a variety of target tasks through transfer learning. To do so, we utilize (1) the encoder for target classification tasks by appending a target task-specific classification head, and (2) the encoder and decoder for target segmentation tasks by replacing the last layer with a 1$\times$1$\times$1 convolutional layer. In this study, we have evaluated the generalization and transferability of TransVW by fine-tuning all the parameters of target models on seven diverse target tasks, including  image  classification and  segmentation  tasks  in  both  2D  and  3D. As summarized in \tablename~\ref{tab:datasets} and detailed in Appendix~\ref{appendix_dataset}, these target tasks offer the following two advantages:

\begin{itemize}
    \item \textbf{Covering a wide range of diseases, organs, and modalities.} This allows us to verify the add-on capability of TransVW on various 3D target tasks, covering a diverse range of diseases (\eg nodule, embolism, tumor), organs (\eg lung, liver, brain), and modalities (\eg CT and MRI). It also enables us to verify the generalizability of TransVW in not only the target tasks on the same dataset as pre-training (\texttt{NCC} and \texttt{NCS}), but also target tasks with a variety of domain shifts (\texttt{ECC}, \texttt{LCS},  and \texttt{BMS}) in terms of modality, scan regions, or dataset. To our best knowledge, we are among the first to investigate cross-domain self-supervised learning in medical imaging. 
    \item \textbf{Enjoying a sufficient amount of annotation.} This paves the way for conducting annotation reduction experiments to verify the annotation efficiency of TransVW.
\end{itemize}

\begin{figure*}[t]
\centerline{\includegraphics[width=0.9\textwidth]{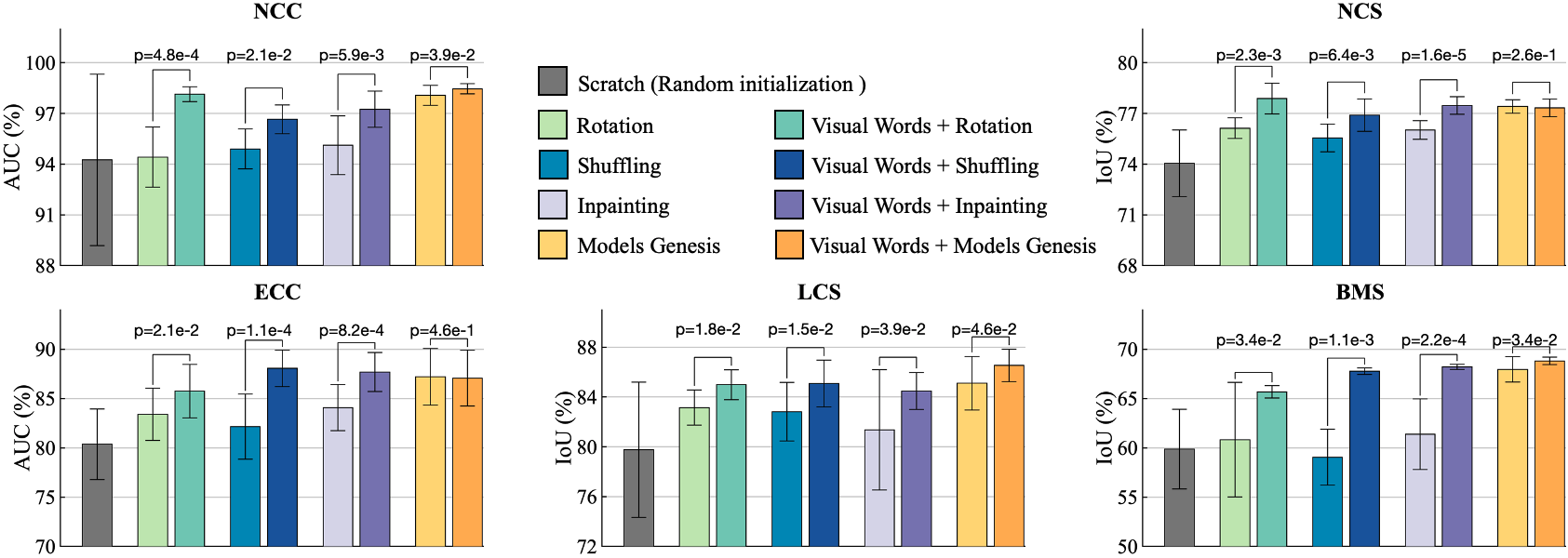}}
\caption{ Our self-supervised learning scheme serves as an add-on, which can be added to enrich  existing self-supervised learning methods. 
By introducing self-discovery and self-classification of visual words, we empower four representative  self-supervised learning advances (\ie Inpainting~\cite{pathak2016context},  Context restoration (Shuffling)~\cite{chen2019self}, Rotation~\cite{gidaris2018unsupervised}, and Models Genesis~\cite{zhou2019models}) to capture more high-level and diverse representations, resulting in substantial ($p<0.05$)  performance improvements on five 3D target  tasks.}
\label{fig:w_wo_semantics}
\end{figure*}

\subsection{Benchmarking TransVW} 
\label{sec:baselines_and_implementation}

For a thorough evaluation, in addition to the training from scratch (the lower-bound baseline), TransVW is compared with a whole range of transfer learning baselines, including both self-supervised and (fully-)supervised methods.

\smallskip
\noindent\textbf{Self-supervised baselines:}  
We compare TransVW with Models Genesis~\cite{zhou2019models}, the state-of-the-art self-supervised learning method for 3D medical imaging, as well as Rubik's cube~\cite{Zhuang2019Self}, the most recent multi-task self-supervised learning method for 3D medical imaging. 
Since most self-supervised learning methods are initially proposed in the context of 2D images, we also have extended three most representative ones~\cite{chen2019self},\cite{pathak2016context},\cite{gidaris2018unsupervised} into their 3D version for a fair comparison.

\smallskip
\noindent\textbf{Supervised baselines:}  
We also examine publicly available {\em fully-supervised} pre-trained models for 3D transfer learning in medical imaging, including NiftyNet~\cite{gibson2018niftynet} and MedicalNet~\cite{chen2019med3d}.  Moreover, we fine-tune Inflated 3D (I3D)~\cite{carreira2017quo} in our 3D target tasks since it has been successfully utilized by Ardila \etal~\cite{ardila2019end} to initialize 3D models for lung nodule detection. 

We utilize 3D U-Net\footnote{\label{foot:3dunet}3D U-Net: \href{https://github.com/ellisdg/3DUnetCNN}{github.com/ellisdg/3DUnetCNN}} for 3D applications, and 
U-Net\footnote{\label{foot:densenet121}Segmentation Models: \href{https://github.com/qubvel/segmentation_models}{github.com/qubvel/segmentation\_models}} with  ResNet-18 as the backbone  for 2D applications. For our pretext task, we modify those architectures by appending fully-connected layers to the end of the encoders for the classification head.
In pretext tasks, we set the  weights of losses as $\lambda_{rec}=1$ and $\lambda_{cls}=0.01$.  All the pretext tasks are trained using Adam optimizer, with a learning rate of 0.001, where $\beta_1 = 0.9$ and $\beta_2 = 0.999$. Regular data augmentation techniques including random flipping, transposing, rotating, elastic transformation, and adding Gaussian noise are utilized in target tasks.  We use the early-stop technique on the validation set
to prevent over-fitting. We run each method ten times on each target task and report the average, standard deviation, and further provide statistical analyses based on independent two-sample \textit{t}-test.

\section{Results}
\label{sec:results}
 This section presents the cornerstones of our results, demonstrating the significance of our self-supervised learning framework. 
  We first integrate our two novel components, self-discovery and self-classification of visual words, into four popular self-supervised methods~\cite{gidaris2018unsupervised,pathak2016context,chen2019self,zhou2019models},  suggesting that these two components can be  adopted  as an add-on to  enhance the existing self-supervised methods. We then compare our TransVW with the current state-of-the-art approaches in a triplet of aspects:  transfer learning performance, convergence speedup, and annotation cost reduction, concluding that TransVW is an annotation-efficient method for medical image analysis. 

\subsection{ TransVW is an add-on scheme}
\label{sec:results_addon}

Our self-discovery  and  self-classification components can readily serve as an add-on to enrich existing self-supervised learning approaches. In fact, by  introducing  these two components,  we open the door for the existing image-based self-supervision approaches
to capture a more high-level and diverse visual representation
that reduces the gap between pre-training and semantic transfer learning tasks.

\smallskip
\noindent\textbf{Experimental setup:} To study the add-on capability of our self-discovery and self-classification components, we incorporate them into four representative self-supervised  methods, including  (1) Models Genesis~\cite{zhou2019models}, which restores the original image patches from the transformed ones; (2) Inpainting~\cite{pathak2016context}, which predicts the missing parts of the  input images; (3) Context restoration~\cite{chen2019self}, which restores the original images from the distorted ones that are obtained by shuffling small patches within the images; and (4) Rotation~\cite{gidaris2018unsupervised}, which predicts the rotation angles that are applied to the input images.
Since all the reconstruction-based self-supervised methods under study~\cite{zhou2019models,pathak2016context,chen2019self} utilize encoder-decoder architecture with skip connections in between, we append additional fully-connected layers to the end of the encoder, enabling models to learn image representation simultaneously from  classification and restoration tasks. 
For Rotation, the network only includes an encoder, followed by two classification heads to learn representations from rotation angle prediction as well as the visual words classification tasks. 
Note that the original self-supervised methods in~\cite{gidaris2018unsupervised},~\cite{pathak2016context}, and~\cite{chen2019self} were implemented in 2D, but we have extended them into 3D.

\begin{table*}[t]
\begin{center}
\begin{threeparttable}
\caption{
 TransVW significantly outperforms training from scratch, and achieves the best or comparable performance in five 3D target applications over five self-supervised and three  publicly available supervised pre-trained 3D models.  
We evaluate the classification (\ie \texttt{NCC} and  \texttt{ECC}) and segmentation (\ie \texttt{NCS}, \texttt{LCS}, and \texttt{BMS}) target tasks under  AUC and IoU metrics, respectively. For each target task, we show the average performance and standard deviation across ten runs, and further perform  independent two sample t-test between the best (bolded) vs. others and
highlighted boxes in green when they are not statistically significantly different at $p = 0.05$ level.
}
\label{tab:3d_transvw_top}

\begin{tabular*}{0.99\textwidth}
{p{0.15\linewidth} P{0.03\linewidth} P{0.161\linewidth} P{0.00001\linewidth} | P{0.00001\linewidth}
p{0.085\linewidth} p{0.085\linewidth} 
p{0.085\linewidth}p{0.085\linewidth}p{0.085\linewidth}}
\hline
\multicolumn{3}{c}{Pre-training task} & &&\multicolumn{5}{c}{Target tasks}\\
\cline{1-3} \cline{6-10} Method & Supervised & Dataset & & &
\centering \texttt{NCC} (\%) & \centering \texttt{NCS}$^1$ (\%) & \centering \texttt{ECC}$^2$ (\%) & \centering \texttt{LCS}$^3$ (\%)  &  \texttt{BMS}$^4$(\%)\\
\hline
Random & \xmark & N/A & & &94.25$\pm$5.07 &74.05$\pm$1.97 & 80.36$\pm$3.58 & 79.76$\pm$5.42 & 59.87$\pm$4.04\\
\hline
NiftyNet~\cite{gibson2018niftynet} & \cmark& Pancreas-CT, BTCV~\cite{gibson2018automatic} &&& 94.14$\pm$4.57 & 52.98$\pm$2.05 & 77.33$\pm$8.05 & 83.23$\pm$1.05 & 60.78$\pm$1.60\\
MedicalNet~\cite{chen2019med3d}& \cmark & 3DSeg-8~\cite{chen2019med3d} &&&95.80$\pm$0.51 & 75.68$\pm$0.32 & 86.43$\pm$1.44 & 85.52$\pm$0.58 & 66.09$\pm$1.35\\
I3D~\cite{carreira2017quo} &\cmark & Kinetics~\cite{carreira2017quo} &&& \cellcolor{maroon!15} 98.26$\pm$0.27 & 71.58$\pm$0.55 & 80.55$\pm$1.11 & 70.65$\pm$4.26 & 67.83$\pm$0.75\\
\hline
Inpainting~\cite{pathak2016context} & \xmark &LUNA16~\cite{setio2017validation} &&& 95.12$\pm$1.74 & 76.02$\pm$0.55 & 84.08$\pm$2.34 & 81.36$\pm$4.83 & 61.38$\pm$3.84\\
Context restoration~\cite{chen2019self} & \xmark &LUNA16~\cite{setio2017validation} &&& 94.90$\pm$1.18 & 75.55$\pm$0.82 & 82.15$\pm$3.3 & 82.82$\pm$2.35 & 59.05$\pm$2.83 \\
Rotation~\cite{gidaris2018unsupervised} & \xmark &LUNA16~\cite{setio2017validation} &&& 94.42$\pm$1.78 & 76.13$\pm$.61 & 83.40$\pm$2.71 & 83.15$\pm$1.41 & 60.53$\pm$5.22\\
Rubik's Cube~\cite{Zhuang2019Self} & \xmark &LUNA16~\cite{setio2017validation} &&& 96.24$\pm$1.27 & 72.87$\pm$0.16 &  80.49$\pm$4.64 &  75.59$\pm$0.20 & 62.75$\pm$1.93\\
Models Genesis~\cite{zhou2019models} & \xmark &LUNA16~\cite{setio2017validation} &&& 98.07$\pm$0.59 & \cellcolor{maroon!15} 77.41$\pm$0.40 & \cellcolor{maroon!15} \textbf{87.2$\pm$2.87} & 85.1$\pm$2.15 & 67.96$\pm$1.29\\
\hline
VW classification  & \xmark &LUNA16~\cite{setio2017validation} &&& 97.49$\pm$0.45 & 76.93$\pm$0.87 & 84.25$\pm$3.91 & 84.14$\pm$1.78 & 64.02$\pm$0.98\\
VW restoration  & \xmark &LUNA16~\cite{setio2017validation} &&& 98.10$\pm$0.19 & \cellcolor{maroon!15} \textbf{77.70$\pm$0.59} & \cellcolor{maroon!15} 86.20$\pm$3.21 & 84.57$\pm$2.20 & 67.78$\pm$0.57\\
TransVW & \xmark & LUNA16~\cite{setio2017validation} &&& \cellcolor{maroon!15} \textbf{98.46$\pm$0.30} & \cellcolor{maroon!15} 77.33$\pm$0.52 &  \cellcolor{maroon!15} 87.07$\pm$2.83 & \cellcolor{maroon!15} \textbf{86.53$\pm$1.30} & \cellcolor{maroon!15} \textbf{68.82$\pm$0.38}\\
\hline
\end{tabular*}

\begin{tablenotes}
       \item  $^1$   \cite{wu2018joint} holds a Dice of 74.05\% vs. $75.85\pm 0.83\%$ (ours)
       \item  $^2$  \cite{zhou2017fine} holds an AUC of 87.06\% vs. 87.07\%$\pm$2.83\% (ours)
       \item  $^3$   \cite{isensee2020automated} holds a Dice of $95.76\%$ vs.  $95.84\%\pm0.07\%$ (ours using nnU-Net framework) 
        \item $^4$  MR Flair images are only utilized for segmenting brain tumors, so the results are not submitted to BraTS-2018. 
    \end{tablenotes}
    \end{threeparttable}
    \end{center}
\end{table*}

\begin{figure*}[!t]
\centerline{\includegraphics[width=\textwidth]{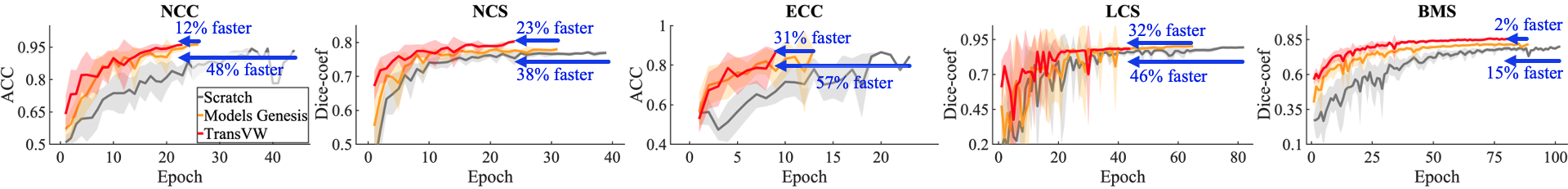}}
\caption{Fine-tuning from TransVW provides better optimization and accelerates the training process in comparison with training from scratch as well as state-of-the-art Models Genesis~\cite{zhou2019models}, as demonstrated by the learning curves for the five 3D target tasks. All models are evaluated on the validation set, and the average accuracy and dice-coefficient over ten runs are plotted for the classification and segmentation tasks, respectively.}
\label{fig:learning_curve}
\end{figure*}

\noindent\textbf{Observations:} Our results in \figurename~\ref{fig:w_wo_semantics} demonstrate that incorporating visual words with existing self-supervised methods consistently improves their performance across five 3D target tasks.
Specifically, visual words significantly improve Rotation by 3\%, 1.5\%, 2\%, 1.5\%, and 5\%; Context restoration by 1.75\%, 1\%, 5\%, 2\%, 
and 8\%; Inpainting by 2\%, 1.5\%, 3\%, 3\%, and 6\% in \texttt{NCC}, \texttt{NCS}, \texttt{ECC}, \texttt{LCS}, and \texttt{BMS} applications, respectively. Moreover, visual words significantly advance Models Genesis by 1\%, 1.5\%, and 1\% in \texttt{NCC}, \texttt{LCS}, and \texttt{BMS}, respectively.

\noindent\textbf{Discussion: How can TransVW improve representation learning?}
Most existing self-supervised learning methods, such as  predicting contexts ~\cite{pathak2016context,chen2019self,zhou2019models,Vincent2008Extracting} or discriminating image transformations~\cite{gidaris2018unsupervised,noroozi2016unsupervised,Zhuang2019Self,Jenni2020Steering}, concentrate on learning the visual representation of each image individually, thereby, overlooking the notion of semantic similarities across different images~\cite{Zhan2020Online,chaitanya2020contrastive}. In contrast, appreciating the recurrent anatomical structure in medical images, our self-discovery can extract meaningful visual words across different images and assign unique semantic pseudo labels to them. By classifying the resultant visual words according to their pseudo labels, our self-classification is enforced to explicitly recognize these visual words across different images—grouping similar visual words together while separating dissimilar ones apart. Consequently, integrating our two novel components into existing self-supervised methods empowers the model to not only learn the local context within a single image but also learn the semantic similarities of the consistent and recurring visual words across images. It is worth noting that self-discovery and self-classification should be considered a significant add-on in terms of methodology. As shown in \figurename~\ref{fig:w_wo_semantics}, simply adding these two components on top of four popular self-supervised methods can noticeably improve their fine-tuning performance.

\subsection{TransVW is an annotation-efficient method}
\label{sec:results_annotation_efficient}

To ensure  that our method addresses annotation scarcity challenge in medical imaging, we first precisely define the annotation-efficiency term.  We consider a method as \textbf{ annotation-efficient} if (1) it achieves superior performance using the same amount of annotated training data, (2) it reduces the training time using the same amount of annotated data, or (3) it offers the same performance but requires less annotated training data.
Based on this definition, we adopt a rigorous \textbf{three-pronged} approach in evaluating our work, by considering not only the transfer learning performance but also the acceleration of the training process and label efficiency on a variety of target tasks,
demonstrating that TransVW is an annotation-efficient method.

\subsubsection {TransVW provides superior transfer learning performance}
\label{sec:result_annot_effic_performance}
A generic pre-trained model transfers well to many different target tasks, indicated by considerable performance improvements. Thus, we first evaluate the generalizability of TransVW in terms of improving the performance of various medical tasks across diseases, organs, and modalities. 

\smallskip
\noindent\textbf{Experimental setup:} We fine-tune TransVW on five 3D target tasks, as described in \tablename~\ref{tab:datasets}, covering classification and segmentation. We investigate the generalizability of TransVW  not only in the target tasks with the pre-training dataset   (\texttt{NCC} and \texttt{NCS}), but also in the target tasks with a variety of domain shifts (\texttt{ECC}, \texttt{LCS}, and \texttt{BMS}).

\smallskip
\noindent\textbf{Observations:} Our evaluations in \tablename~\ref{tab:3d_transvw_top} suggest three major results. 
Firstly, TransVW significantly outperforms training from scratch in all five target tasks under study by  a large margin and also stabilizes the overall performance.

Secondly, TransVW surpasses all self-supervised counterparts in the five target tasks. Specifically, TransVW significantly outperforms Models Genesis, state-of-the-art self-supervised 3D models pre-trained using image restoration, in three applications, \ie \texttt{NCC}, \texttt{LCS}, and \texttt{BMS}, and offers equivalent performance in \texttt{NCS} and \texttt{ECC}.  Moreover, TransVW yields remarkable improvements over Rubik's cube, the most recent 3D multi-task self-supervised method, in all five applications. Particularly,  Rubik's cube formulates a multi-task learning objective solely based on contextual cues within single images while TransVW benefits from semantic supervision of anatomical visual words, resulting in more enhanced representations.

 Thirdly, TransVW achieves superior performance in comparison with publicly available fully-supervised  pre-trained 3D models, \ie NiftyNet, MedicalNet, and I3D, in all five target tasks.  It is noteworthy that our TransVW does not  solely depend on  the architecture capacity to achieve the best performance since it has much fewer  model parameters than its  counterparts. Specifically, TransVW is trained on basic 3D U-Net with  23M parameters, while MedicalNet with ResNet-101 as the backbone (reported as the best performing model in~\cite{chen2019med3d}) carries 85.75M parameters, and I3D contains 25.35M parameters in the encoder. Although NiftyNet model is offered with 2.6M parameters, its performance is not as good as its supervised counterparts  in any of the target tasks.

\begin{figure*}[t]
\centerline{\includegraphics[width=\textwidth]{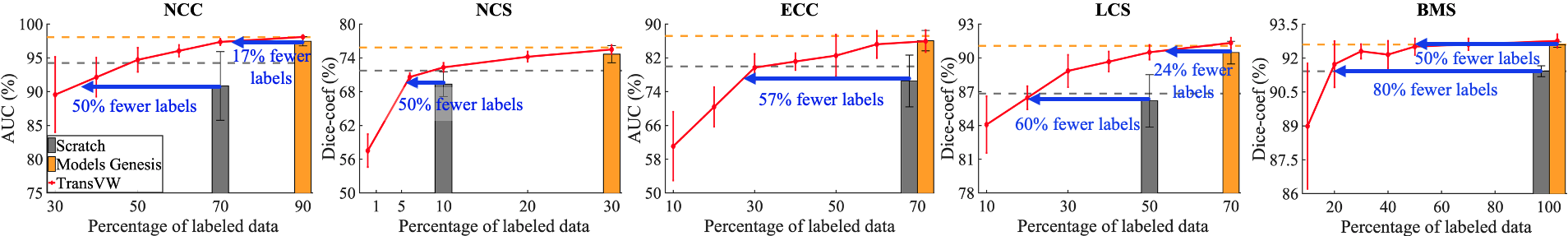}}
\caption{
Fine-tuning  TransVW reduces the annotation cost by 50\%, 50\%, 57\%, 60\%, and 80\%  in \texttt{NCC}, \texttt{NCS}, \texttt{ECC}, \texttt{LCS}, and \texttt{BMS} applications, respectively, when comparing with training from scratch. Moreover,  TransVW reduces the annotation efforts by 17\%, 24\%, and 50\% in \texttt{NCC}, \texttt{LCS}, and \texttt{BMS} applications, respectively, compared with state-of-the-art Models Genesis~\cite{zhou2019models}.
The horizontal gray and orange lines show the performance achieved by training from scratch and Models Genesis, respectively, when using the entire  training data. The gray and orange bars indicate the minimum portion of training data that is required for training models from scratch and Models Genesis to achieve the comparable performance (based on the statistical analyses) with the corresponding models when training with the entire  training data.  
}
\label{fig:annotation_saving}
\end{figure*} 
 Interestingly, although TransVW is pre-trained on chest CT scans, it is still beneficial for  different organs, diseases, datasets, and even modalities. In particular,  the pulmonary embolism false positive reduction (\texttt{ECC}) is on Contrast-Enhanced CT scans, which may appear differently from the normal CT scans that are used for pre-training; yet, according to \tablename~\ref{tab:3d_transvw_top}, TransVW obtains a 7\% improvement over training 3D models from scratch in this task. Additionally, fine-tuning from TransVW provides a substantial gain in liver segmentation (\texttt{LCS}) accuracy despite the noticeable differences between pretext and target domains in terms of organs (lung vs. liver) and datasets (LUNA 2016 vs. LiTS 2017). We further examine the transferability of TransVW in brain tumor segmentation on MRI Flair images (\texttt{BMS}). Referring to   \tablename~\ref{tab:3d_transvw_top}, despite the marked differences in organs, datasets, and even modalities between the pretext and \texttt{BMS} target task, we still observe a significant performance boost from fine-tuning TransVW  in comparison with learning   from scratch.
Moreover, TransVW significantly outperforms Models Genesis in the most distant target domains  from the pretext task, \ie  \texttt{LCS}, and \texttt{BMS}. 

\smallskip
\noindent\textbf{Discussion: How can TransVW improve the performance of cross-domain target tasks?}
Learning universal representations that can transfer effectively to a wide range of target tasks is one of the supreme goals  of  computer  vision  in  medical  imaging.  The  best  known  existing  examples  of  such  representations  are  pre-trained  models  on  ImageNet  dataset.  Although  there  are  marked  differences  between  natural  and  medical  images, the image representations learned from ImageNet can be beneficial not only for natural imaging~\cite{Kornblith2019Do} but also for  medical  imaging  ~\cite{tajbakhsh2016convolutional,Neyshabur2020What}.  Therefore,  rather  than  developing  pre-trained  models  specifically  for  a  particular dataset/organ,  TransVW  aims  to  develop  generic  pre-trained  3D  models  for  medical  image  analysis  that  are  not biased to idiosyncrasies of the pre-training task and dataset and generalize effectively across organs, datasets, and modalities.

As is well-known, CNNs trained on large scale visual data form feature hierarchies; lower layers of deep networks are in charge of general features while higher layers contain more specialized features for target domains\cite{yosinski2014transferable,raghu2019transfusion,Neyshabur2020What}. Due to generalizability of low and mid-level features, they can lead to significant benefits of transfer learning, even when there is a substantial domain gap between the pretext and  target  task~\cite{Neyshabur2020What, raghu2019transfusion}.  Therefore,  while  TransVW  is  pre-trained  solely  on  chest  CT  scans,  it  still  elevates  the target task performance in different organs, datasets, and modalities since its low and mid-level features can be reused in the target tasks.

\subsubsection{TransVW accelerates the training process}
\label{sec:result_annot_effic_training_time}
Although accelerating the training of deep neural networks is arguably an influential line of research~\cite{wang2019e2train}, its importance is often underappreciated in the medical imaging literature. Transfer learning provides a warm-up initialization that enables target models to converge faster and mitigates the vanishing and exploding gradient problems. 
In that respect, we  argue that a good pre-trained model should  yield better target task performance  with  less training time. Hence, we further evaluate TransVW in terms of accelerating the training process of various medical tasks. 

\smallskip
\noindent\textbf{Experimental setup:} We compare the convergence speedups obtained by TransVW 
with training from scratch (the lower bound baseline) and Models Genesis (the state-of-the-art baseline) in our five 3D target tasks.  For conducting fair comparisons in all experiments,  all methods benefit from the same data augmentation and use the same  network architecture while we endeavor to optimize each model with the best-performing hyper-parameters. 

\smallskip
\noindent\textbf{Observations:} 
\figurename~\ref{fig:learning_curve} presents the learning curves for training from scratch, fine-tuning from Models Genesis, and fine-tuning from TransVW.  Our results demonstrate that initializing 3D models from TransVW  remarkably accelerates the training process of target models in comparison with not only learning from scratch but also Models Genesis in all five target tasks.  
These results imply that  
 TransVW captures  representations  that are more aligned with the subsequent target tasks, leading to faster convergence of the target models. 
 
Putting the transfer learning performance on five target tasks in \tablename~\ref{tab:3d_transvw_top} and the training times in \figurename~\ref{fig:learning_curve} together,  TransVW demonstrates  significantly  better  or  equivalent  performance  with  remarkably  less  training  time  in  comparison  to  its  3D counterparts.
Specifically, TransVW significantly outperforms Models Genesis in terms of both performance and saving training time in three out of five applications, \ie \texttt{NCC}, \texttt{LCS}, and \texttt{BMS}, and achieves equivalent performance in  \texttt{NCS} and \texttt{ECC} but in remarkably less time. 
 Altogether,  we believe that TransVW can serve as a primary source of transfer learning for 3D medical imaging applications to boost the performance and accelerate the training of target tasks.

\subsubsection{TransVW reduces the annotation cost} 
\label{sec:result_annot_effic_annot_cost}
Transfer learning  yields more accurate models by reusing the previously learned knowledge in target tasks with limited annotations.  
This is because a good representation should not need many samples to learn about a concept~\cite{goyal2019scaling}. 
Thereby, we conduct experiments on partially labeled data to  investigate transferability of TransVW in small data regimes.

\smallskip
\noindent\textbf{Experimental setup:} We compare the transfer learning performance of TransVW using partial labeled data with training from scratch and fine-tuning from  Models Genesis in five 3D target tasks. For clarity, we determine the minimum required data for training from scratch and fine-tuning Models Genesis to meet the comparable performance (based on independent two-sample \textit{t}-test) when  training using  the entire training data. Moreover, we investigate the minimum required data for TransVW to meet the equivalent performance that  training from scratch and fine-tuning Models Genesis can achieve.

\smallskip
\noindent\textbf{Observations:}
\figurename~\ref{fig:annotation_saving} illustrates the results of using the partial amount of labeled data during the training of five 3D target tasks. As an illustrative example, in lung nodule false positive reduction (\texttt{NCC}), our results demonstrate that using only 35\% of training data, TransVW achieves equivalent performance to training from scratch using 70\% of data. Therefore, around 50\% of the annotation cost in \texttt{NCC} can be reduced by fine-tuning models from TransVW compared with training from scratch. In comparison with Models Genesis in the same application (\texttt{NCC}), TransVW with 75\% of data achieves equal performance with Models Genesis using 90\% of data. Therefore, about 17\% of the annotation cost associated with fine-tuning from Models Genesis in \texttt{NCC} is recovered by fine-tuning from TransVW. In general, transfer learning from TransVW reduces the annotation cost by 50\%, 50\%, 57\%, 60\%, and 80\% in comparison with training from scratch in \texttt{NCC}, \texttt{NCS}, \texttt{ECC}, \texttt{LCS}, and \texttt{BMS} applications, respectively. In comparison with Models Genesis, TransVW reduces the annotation efforts by 17\%, 24\%, and 50\% in \texttt{NCC}, \texttt{LCS}, and \texttt{BMS} applications, respectively, and both models performs equally in the \texttt{NCS} and \texttt{ECC} applications. 
These results suggest that TransVW achieves state-of-the-art or comparable performance over other self-supervised  approaches while being  more efficient, \ie less annotated data is required for training high-performance target models.

\smallskip
\noindent\textbf{Summary:} We demonstrate that TransVW  provides more generic and transferable representations compared with self-supervised and supervised 3D competitors, confirmed by  our  evaluations on a triplet of transfer learning performance, optimization speedup, and annotation cost. 
To further illustrate the effectiveness of our framework, we adopt TransVW to the nnU-Net~\cite{isensee2020automated},  a state-of-the-art segmentation framework in  medical  imaging, and evaluate  it  on  liver  tumor  segmentation  task from the Medical Segmentation Decathlon~\cite{Amber2019large}. Our results demonstrate that TransVW obtains improvements in segmentation accuracy over training from scratch and Models Genesis by 2.5\% and 1\%,  respectively (details in Appendix~\ref{appendix_discussion_nnunet}). These results, in line with our previous results, reinforce our main insight that TransVW provides an annotation-efficient solution for 3D medical imaging.

 \begin{figure*}[t]
\centerline{\includegraphics[width=\textwidth]{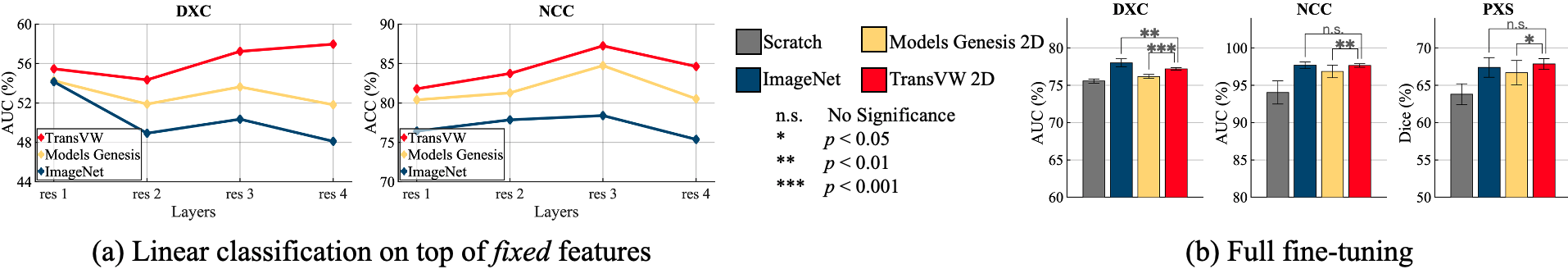}}
\caption{We compare the learned representation of TransVW 2D with Models Genesis 2D (self-supervised) and ImageNet (fully-supervised) by (a) training linear classifiers on top of fixed features, and (b) full fine-tuning of the models on 2D applications. In the linear evaluations (a), TransVW representations are transferred better across all the layers on both \texttt{DXC} and \texttt{NCC} in comparison with Models Genesis 2D and ImageNet, demonstrating more generalizable features. Based on the fine-tuning results (b), TransVW 2D significantly surpasses training from scratch and Models Genesis 2D, and achieves equivalent performance with ImageNet in \texttt{NCC} and \texttt{PXS}.   }
\label{fig:2d_linear_and_finetuning}
\end{figure*}

\section{Ablation experiments}
In this section, we first conduct ablation experiments to illustrate the contribution of different components to the performance of TransVW. We then train a 2D model using chest X-ray images, called TransVW 2D, and compare it with state-of-the-art 2D models. In the 2D experiments, we consider three target tasks: thorax diseases classification (\texttt{DXC}), pneumothorax segmentation (\texttt{PXS}), and lung nodule false positive reduction (\texttt{NCC}). We evaluate \texttt{NCC} in a 2D slice-based solution, where the 2D representation is obtained by extracting axial slices from the volumetric dataset.

\subsection{Comparing individual self-supervised tasks}
\label{sec:abl_isolated}
Our TransVW takes the advantages of two sources in representation learning: self-classification and self-restoration of visual words. Therefore,  we first directly compare the two isolated tasks and then investigate  whether joint-task learning in TransVW produces more transferable features compared with isolated training schemes. 
In \tablename~\ref{tab:3d_transvw_top}, the last three rows show the transfer learning results of TransVW and each of the  individual tasks in five 3D applications.  
 According to  the statistical analysis results, self-restoration and self-classification reveal no significant difference ($p$-value $>$ 0.05) in three target tasks,  \texttt{NCS}, \texttt{ECC}, and \texttt{LCS}, and self-restoration achieves significantly better performance in \texttt{NCC} and \texttt{BMS}.
Despite the success of self-restoration in encoding fine-grained anatomical information from individual visual words, it neglects the semantic relationships across different visual words. In contrast, our novel self-classification component explicitly encodes the semantic similarities that presents across visual  words into the learned embedding space. Therefore, as evidenced by \tablename~\ref{tab:3d_transvw_top}, integrating self-classification and self-restoration into a single framework yields a more comprehensive representation that can guarantee the highest target task performance. In particular, TransVW outperforms each isolated task in four applications, \ie \texttt{NCC}, \texttt{ECC}, \texttt{LCC}, and \texttt{BMS}, and provides comparable performance with self-restoration in \texttt{NCS}.

\subsection{Evaluating 2D applications}
\label{sec:ablation_2d}

We evaluate our TransVW 2D with Models Genesis 2D (self-supervised) and ImageNet (fully-supervised) pre-trained models in two experimental settings: (1) linear evaluation on top of the fixed features from the pre-trained network, and (2) full fine-tuning of the pre-trained network for target tasks. 

\smallskip
 \noindent\textbf{Linear evaluation:} To evaluate the quality of the learned representations, we follow the common practice in \cite{Zhan2020Online,goyal2019scaling}, which trains linear classifiers on top of the fixed features obtained from  various layers of  the pre-trained networks. Specifically, for ResNet-18 backbone, we extract image features from the last layer of every  residual  stage (denoted as res1, res2, etc)~\cite{goyal2019scaling}, and then  evaluate  them in two classification target tasks (\texttt{DXC} and  \texttt{NCC}).
Based on our results in \figurename~\ref{fig:2d_linear_and_finetuning}a, TransVW 2D representations are transferred better across all the layers on both target tasks in comparison with Models Genesis 2D and ImageNet, demonstrating the generalizability of TransVW 2D representations. Specifically, in thorax diseases classification (\texttt{DXC}), which is in the same dataset as the pretext task, the best performing features are extracted from  res4  in the last layer of the TransVW 2D network. This indicates that  TransVW 2D encourages the models to squeeze out high-level representations, which are aligned with the target task, in the deeper layers of the network. Moreover, in lung nodule false positive reduction (\texttt{NCC}), which presents a domain shift compared with the pretext task, TransVW 2D remarkably reduces the performance gap between res3 and res4 features compared with  Models Genesis 2D and ImageNet. This suggests that TransVW reduces the overfitting of res4 features to the pretext task and dataset, resulting in more generic features.

\smallskip
\noindent\textbf{Full fine-tuning:} We evaluate the initialization provided by our TransVW 2D via fine-tuning it for three  2D target  tasks,  covering  classification  (\texttt{DXC} and \texttt{NCC})  and  segmentation  (\texttt{PXS})  in  X-ray  and  CT.
As evidenced by our statistical analysis in \figurename~\ref{fig:2d_linear_and_finetuning}b,  TransVW 2D: (1) significantly surpasses training from scratch and Models Genesis 2D in all three applications, and (2) achieves equivalent performance with ImageNet in \texttt{NCC} and \texttt{PXS}, which is a significant achievement because to date, all self-supervised approaches lag behind fully supervised training~\cite{hendrycks2019using,caron2019unsupervised,zhang2019aet}. Taken together, these results indicate that  our TransVW 2D, which comes at zero annotation cost, generalizes well across tasks, datasets, and modalities.  
 
\smallskip
\noindent\textbf{Discussion: Is there any correspondence between linear evaluation and fine-tuning performance?}  As seen in \figurename~\ref{fig:2d_linear_and_finetuning},  ImageNet models underperform in linear evaluations with fixed features; however,  full fine-tuning of ImageNet features yield higher (in \texttt{DXC}) or equal  (in \texttt{NCC} and \texttt{PXS}) performance compared with  TransVW 2D. 
We surmise that although ImageNet models leverage large-scale annotated data during pre-training, due to the marked domain gap between medical and natural images, the fixed features of ImageNet models may not be aligned with medical applications. However, transfer learning from ImageNet models can still provide a good initialization point for the CNNs since its low and mid-level features can be reused in the target tasks. Thus, fine-tuning their features on a large-scale dataset such as ChestX-ray14 could mitigate the discrepancy between natural and medical domains, yielding good target task performance.

  Our observations in line with~\cite{Newell2020How} suggest that although linear evaluation is informative for utilizing fixed features, it may not have a strong correlation to the fine-tuning performance. 

\section{Related works}
\label{sec:related_works}

\noindent\textbf{Bag-of-Visual-Words (BoVW):}  
The  BoVW  model~\cite{Sivic2003Video} represents images by local invariant features~\cite{Lowe2004Distinctive} that are condensed into a single vector representation.
BoVW and its extensions  have been widely used in various
tasks~\cite{Ng2015Exploiting}. 
A major drawback of BoVW is that the extracted visual words cannot be transferred and fine-tuned for new tasks and datasets like CNN models.
To address this challenge, recently, a few works have integrated BoVW in the training pipeline of  CNNs~\cite{Gidaris2020Learning,Arandjelovic2016NetVLAD}. Among them,  Gidaris \etal~\cite{Gidaris2020Learning}  proposed a  pretext task based on the BoVW pipeline, which first discretizes images to a set of spatially dense visual words using a pre-trained network, and then  trains  a  second  network  to  predict  the BoVW histogram of the images. While this approach displays impressive results for natural images, the extracted visual words may not be intuitive and explainable from a medical perspective since they are automatically determined in feature space. Moreover, using K-means clustering in creating the visual vocabulary may lead to imbalanced clusters of visual words (known as cluster degeneracy)~\cite{gansbeke2020scan}. Our method differs from previous works in  (1)  automatically discovering visual words that carry explainable semantic information from medical perspective,
(2) bypassing the clustering, reducing the training time and leading to the balanced classes of visual words, and (3) proposing a novel pretext rather than predicting the BoVW histogram.  

\smallskip
\noindent\textbf{Self-supervised learning:}  
Self-supervised learning methods aim to learn general representations from unlabeled data. In this paradigm, a neural network is trained on a manually designed (pretext) task for which ground-truth is available for free. The learned representations can be later fine-tuned on numerous target tasks with limited annotated data.
A broad variety of self-supervised methods have been proposed for pre-training CNNs in natural images domain,  solving jigsaw puzzles~\cite{noroozi2016unsupervised}, predicting image rotations~\cite{gidaris2018unsupervised}, inpainting of missing parts~\cite{pathak2016context}, clustering images and then predicting the cluster assignments~\cite{Zhan2020Online,caron2018deep}, and   removing noise from noisy images~\cite{Vincent2008Extracting}.
However, self-supervised learning is a relatively new trend in medical imaging. Recent methods, including 
colorization of colonoscopy images~\cite{ross2018exploiting},  anatomical positions prediction within cardiac MR images~\cite{Bai2019Self}, context restoration~\cite{chen2019self}, and Rubik's cube recovery~\cite{Zhuang2019Self}, which were developed individually for specific target tasks, have a limited generalization ability over multiple tasks. 
TransVW distinguishes itself from all other existing works by explicitly employing the strong yet free semantic supervision signals of visual words, leading to a generic pre-trained model effective for various target tasks. 
Recently, Zhou \etal~\cite{zhou2019models} proposed four effective image transformations for learning generic autodidactic models through a restoration-based task for 3D medical imaging. While our method derives the transformations from Models Genesis~\cite{zhou2019models}, it shows \textbf{three significant advancements}. First, Models Genesis has only one self-restoration component, while we introduce two more novel components: self-discovery and self-classification, which are sole factors in the performance gain. Second, our method learns  semantic representation from the consistent and recurring visual words discovered during our self-discovery phase, but Models Genesis learns representation from random sub-volumes with no semantics, since no semantics can be discovered from random sub-volumes. Finally,  our method serves as an add-on for boosting other self-supervised methods while Models Genesis do not offer such advantage.

\smallskip
\noindent\textbf{Our previous work:} Haghighi \etal~\cite{haghighi2020learning} first proposed to utilize consistent anatomical patterns for training semantics-enriched pre-trained models.  The current work presents several extensions to the preliminary
version: (1) We introduce a new concept: \textbf{transferable visual words}, where the recurrent anatomical structures in medical images are anatomical visual words, which can be automatically discovered from unlabeled medical images, serving as strong yet free supervision signals for training deep models; 
(2) We extensively investigate the  add-on  capability of our self-discovery and self-classification, demonstrating that they can boost existing self-supervised learning methods (\sectionname~\ref{sec:results_addon}); (3) We expand our 3D target tasks by adding  pulmonary embolism false positive reduction, indicating that our TransVW generalizes effectively across  organs,  datasets, and  modalities (\sectionname~\ref{sec:experiments_finetune}); (4) We extend  Rotation~\cite{gidaris2018unsupervised} into its 3D version as an additional baseline (\sectionname~\ref{sec:result_annot_effic_performance});
 (5) We  adopt  a  rigorous  three-pronged  approach in  evaluating the transferability of our  work, including  transfer learning performance, convergence speedups,  and annotation efficiency, highlighting that TransVW is an annotation-efficient solution for 3D medical imaging. As part of this endeavor we illustrate that transfer learning from TransVW provides better optimization and accelerates the training process (\sectionname~\ref{sec:result_annot_effic_training_time}) and dramatically reduces annotation efforts (\sectionname~\ref{sec:result_annot_effic_annot_cost}).
 (6) We conduct linear evaluations on top of the fixed features, showing that TransVW 2D provides more generic representations by reducing the overfitting to the pretext task (\sectionname~\ref{sec:ablation_2d}); and  (7) We conduct  ablation studies on five 3D target tasks to search for an effective number of visual word classes (Appendix~\ref{appendix_abl_num_vws}). 

\section{Conclusion}
A key contribution of ours is designing a self-supervised learning framework that not only allows deep models to learn common visual representation from image data directly, but also leverages the semantics associated with the recurrent anatomical patterns across  medical  images, resulting in generic semantics-enriched image representations. Our extensive experiments  demonstrate the annotation-efficiency of TransVW by offering  higher performance and faster  convergence with reduced annotation cost  in comparison with publicly available 3D models pre-trained by  not only self-supervision but also full supervision. More importantly, TransVW can be used as an add-on scheme to substantially improve other self-supervised methods. We attribute these outstanding results to the compelling deep semantics derived from recurrent visual words resulted from consistent anatomies naturally embedded in medical images.

\section*{Acknowledgment}
This research has been supported partially by ASU and Mayo Clinic through a Seed Grant and an Innovation Grant, and partially by the NIH under Award Number R01HL128785.  The content is solely the responsibility of the authors and does not necessarily represent the official views of the NIH.
 This work has utilized the GPUs provided partially by the ASU Research Computing and partially by the Extreme Science and Engineering Discovery Environment (XSEDE) funded by the National Science Foundation (NSF) under grant number ACI-1548562.
We thank Zuwei Guo for implementing Rubik's cube~\cite{Zhuang2019Self}, Md Mahfuzur Rahman Siddiquee for examining NiftyNet~\cite{gibson2018niftynet}, Jiaxuan Pang for evaluating I3D~\cite{carreira2017quo}, Shivam Bajpai for helping in adopting TransVW to nnU-Net~\cite{isensee2020automated}, and
Shrikar Tatapudi for helping improve the writing of this paper.  The content of this paper is covered by patents pending.


\begin{thebibliography}{10}
\providecommand{\url}[1]{#1}
\csname url@samestyle\endcsname
\providecommand{\newblock}{\relax}
\providecommand{\bibinfo}[2]{#2}
\providecommand{\BIBentrySTDinterwordspacing}{\spaceskip=0pt\relax}
\providecommand{\BIBentryALTinterwordstretchfactor}{4}
\providecommand{\BIBentryALTinterwordspacing}{\spaceskip=\fontdimen2\font plus
\BIBentryALTinterwordstretchfactor\fontdimen3\font minus
  \fontdimen4\font\relax}
\providecommand{\BIBforeignlanguage}[2]{{%
\expandafter\ifx\csname l@#1\endcsname\relax
\typeout{** WARNING: IEEEtran.bst: No hyphenation pattern has been}%
\typeout{** loaded for the language `#1'. Using the pattern for}%
\typeout{** the default language instead.}%
\else
\language=\csname l@#1\endcsname
\fi
#2}}
\providecommand{\BIBdecl}{\relax}
\BIBdecl

\bibitem{Sivic2003Video}
J.~Sivic and A.~Zisserman, ``Video google: a text retrieval approach to object
  matching in videos,'' in \emph{Proceedings Ninth IEEE International
  Conference on Computer Vision}, Oct 2003, pp. 1470--1477 vol.2.

\bibitem{gidaris2018unsupervised}
S.~Gidaris, P.~Singh, and N.~Komodakis, ``Unsupervised representation learning
  by predicting image rotations,'' \emph{arXiv preprint arXiv:1803.07728},
  2018.

\bibitem{pathak2016context}
D.~Pathak, P.~Krahenbuhl, J.~Donahue, T.~Darrell, and A.~A. Efros, ``Context
  encoders: Feature learning by inpainting,'' in \emph{Proceedings of the IEEE
  Conference on Computer Vision and Pattern Recognition}, 2016, pp. 2536--2544.

\bibitem{chen2019self}
L.~Chen, P.~Bentley, K.~Mori, K.~Misawa, M.~Fujiwara, and D.~Rueckert,
  ``Self-supervised learning for medical image analysis using image context
  restoration,'' \emph{Medical image analysis}, vol.~58, p. 101539, 2019.

\bibitem{zhou2019models}
Z.~Zhou, V.~Sodha, M.~M. Rahman~Siddiquee, R.~Feng, N.~Tajbakhsh, M.~B. Gotway,
  and J.~Liang, ``Models genesis: Generic autodidactic models for 3d medical
  image analysis,'' in \emph{Medical Image Computing and Computer Assisted
  Intervention -- MICCAI 2019}.\hskip 1em plus 0.5em minus 0.4em\relax Cham:
  Springer International Publishing, 2019, pp. 384--393.

\bibitem{zhou2021models}
Z.~Zhou, V.~Sodha, J.~Pang, M.~B. Gotway, and J.~Liang, ``Models genesis,''
  \emph{Medical Image Analysis}, vol.~67, p. 101840, 2021.

\bibitem{setio2017validation}
A.~A.~A. Setio, A.~Traverso, T.~De~Bel, M.~S. Berens, C.~van~den Bogaard,
  P.~Cerello, H.~Chen, Q.~Dou, M.~E. Fantacci, B.~Geurts \emph{et~al.},
  ``Validation, comparison, and combination of algorithms for automatic
  detection of pulmonary nodules in computed tomography images: the luna16
  challenge,'' \emph{Medical image analysis}, vol.~42, pp. 1--13, 2017.

\bibitem{armato2011lung}
S.~G. Armato~III, G.~McLennan, L.~Bidaut, M.~F. McNitt-Gray, C.~R. Meyer, A.~P.
  Reeves, B.~Zhao, D.~R. Aberle, C.~I. Henschke, E.~A. Hoffman \emph{et~al.},
  ``The lung image database consortium (lidc) and image database resource
  initiative (idri): a completed reference database of lung nodules on ct
  scans,'' \emph{Medical physics}, vol.~38, pp. 915--931, 2011.

\bibitem{tajbakhsh2015computer}
N.~Tajbakhsh, M.~B. Gotway, and J.~Liang, ``Computer-aided pulmonary embolism
  detection using a novel vessel-aligned multi-planar image representation and
  convolutional neural networks,'' in \emph{International Conference on Medical
  Image Computing and Computer-Assisted Intervention}.\hskip 1em plus 0.5em
  minus 0.4em\relax Springer, 2015, pp. 62--69.

\bibitem{bilic2019liver}
P.~Bilic, P.~F. Christ, E.~Vorontsov, G.~Chlebus, H.~Chen, Q.~Dou, C.-W. Fu,
  X.~Han, P.-A. Heng, J.~Hesser \emph{et~al.}, ``The liver tumor segmentation
  benchmark (lits),'' \emph{arXiv preprint arXiv:1901.04056}, 2019.

\bibitem{bakas2018identifying}
S.~Bakas, M.~Reyes, A.~Jakab, S.~Bauer, M.~Rempfler, A.~Crimi, R.~T. Shinohara,
  C.~Berger, S.~M. Ha, M.~Rozycki \emph{et~al.}, ``Identifying the best machine
  learning algorithms for brain tumor segmentation, progression assessment, and
  overall survival prediction in the brats challenge,'' \emph{arXiv preprint
  arXiv:1811.02629}, 2018.

\bibitem{wang2017chestx}
X.~Wang, Y.~Peng, L.~Lu, Z.~Lu, M.~Bagheri, and R.~M. Summers, ``Chestx-ray8:
  Hospital-scale chest x-ray database and benchmarks on weakly-supervised
  classification and localization of common thorax diseases,'' in
  \emph{Proceedings of the IEEE Conference on Computer Vision and Pattern
  Recognition}, 2017, pp. 2097--2106.

\bibitem{PNEchallenge}
\BIBentryALTinterwordspacing
(2019) Siim-acr pneumothorax segmentation. [Online]. Available:
  \url{https://www.kaggle.com/c/siim-acr-pneumothorax-segmentation/}
\BIBentrySTDinterwordspacing

\bibitem{Zhuang2019Self}
X.~Zhuang, Y.~Li, Y.~Hu, K.~Ma, Y.~Yang, and Y.~Zheng, ``Self-supervised
  feature learning for 3d medical images by playing a rubik's cube,'' in
  \emph{Medical Image Computing and Computer Assisted Intervention -- MICCAI
  2019}, D.~Shen, T.~Liu, T.~M. Peters, L.~H. Staib, C.~Essert, S.~Zhou, P.-T.
  Yap, and A.~Khan, Eds.\hskip 1em plus 0.5em minus 0.4em\relax Cham: Springer
  International Publishing, 2019, pp. 420--428.

\bibitem{carreira2017quo}
J.~Carreira and A.~Zisserman, ``Quo vadis, action recognition? a new model and
  the kinetics dataset,'' in \emph{Proceedings of the IEEE Conference on
  Computer Vision and Pattern Recognition}, 2017, pp. 6299--6308.

\bibitem{gibson2018niftynet}
E.~Gibson, W.~Li, C.~Sudre, L.~Fidon, D.~I. Shakir, G.~Wang, Z.~Eaton-Rosen,
  R.~Gray, T.~Doel, Y.~Hu \emph{et~al.}, ``Niftynet: a deep-learning platform
  for medical imaging,'' \emph{Computer methods and programs in biomedicine},
  vol. 158, pp. 113--122, 2018.

\bibitem{chen2019med3d}
S.~Chen, K.~Ma, and Y.~Zheng, ``Med3d: Transfer learning for 3d medical image
  analysis,'' \emph{arXiv preprint arXiv:1904.00625}, 2019.

\bibitem{Jenni2020Steering}
S.~Jenni, H.~Jin, and P.~Favaro, ``Steering self-supervised feature learning
  beyond local pixel statistics,'' in \emph{Proceedings of the IEEE/CVF
  Conference on Computer Vision and Pattern Recognition (CVPR)}, June 2020.

\bibitem{doersch2015unsupervised}
C.~Doersch, A.~Gupta, and A.~A. Efros, ``Unsupervised visual representation
  learning by context prediction,'' in \emph{Proceedings of the IEEE
  International Conference on Computer Vision}, 2015, pp. 1422--1430.

\bibitem{mundhenk2018improvements}
T.~N. Mundhenk, D.~Ho, and B.~Y. Chen, ``Improvements to context based
  self-supervised learning.'' in \emph{Proceedings of the IEEE Conference on
  Computer Vision and Pattern Recognition}, 2018, pp. 9339--9348.

\bibitem{goyal2019scaling}
P.~Goyal, D.~Mahajan, A.~Gupta, and I.~Misra, ``Scaling and benchmarking
  self-supervised visual representation learning,'' in \emph{Proceedings of the
  IEEE/CVF International Conference on Computer Vision (ICCV)}, 2019.

\bibitem{ardila2019end}
D.~Ardila, A.~P. Kiraly, S.~Bharadwaj, B.~Choi, J.~J. Reicher, L.~Peng, D.~Tse,
  M.~Etemadi, W.~Ye, G.~Corrado \emph{et~al.}, ``End-to-end lung cancer
  screening with three-dimensional deep learning on low-dose chest computed
  tomography,'' \emph{Nature medicine}, vol.~25, pp. 954--961, 2019.

\bibitem{gibson2018automatic}
E.~Gibson, F.~Giganti, Y.~Hu, E.~Bonmati, S.~Bandula, K.~Gurusamy, B.~Davidson,
  S.~P. Pereira, M.~J. Clarkson, and D.~C. Barratt, ``Automatic multi-organ
  segmentation on abdominal ct with dense v-networks,'' \emph{IEEE transactions
  on medical imaging}, vol.~37, no.~8, pp. 1822--1834, 2018.

\bibitem{wu2018joint}
B.~Wu, Z.~Zhou, J.~Wang, and Y.~Wang, ``Joint learning for pulmonary nodule
  segmentation, attributes and malignancy prediction,'' in \emph{2018 IEEE 15th
  International Symposium on Biomedical Imaging (ISBI 2018)}.\hskip 1em plus
  0.5em minus 0.4em\relax IEEE, 2018, pp. 1109--1113.

\bibitem{zhou2017fine}
Z.~Zhou, J.~Shin, L.~Zhang, S.~Gurudu, M.~Gotway, and J.~Liang, ``Fine-tuning
  convolutional neural networks for biomedical image analysis: actively and
  incrementally,'' in \emph{Proceedings of the IEEE Conference on Computer
  Vision and Pattern Recognition}, 2017, pp. 7340--7349.

\bibitem{isensee2020automated}
F.~Isensee, P.~F. Jäger, S.~A.~A. Kohl, J.~Petersen, and K.~H. Maier-Hein,
  ``Automated design of deep learning methods for biomedical image
  segmentation,'' 2020.

\bibitem{Vincent2008Extracting}
P.~Vincent, H.~Larochelle, Y.~Bengio, and P.-A. Manzagol, ``Extracting and
  composing robust features with denoising autoencoders,'' in \emph{Proceedings
  of the 25th International Conference on Machine Learning}, ser. ICML
  ’08.\hskip 1em plus 0.5em minus 0.4em\relax Association for Computing
  Machinery, 2008, p. 1096–1103.

\bibitem{noroozi2016unsupervised}
M.~Noroozi and P.~Favaro, ``Unsupervised learning of visual representations by
  solving jigsaw puzzles,'' in \emph{European Conference on Computer
  Vision}.\hskip 1em plus 0.5em minus 0.4em\relax Springer, 2016, pp. 69--84.

\bibitem{Zhan2020Online}
X.~Zhan, J.~Xie, Z.~Liu, Y.-S. Ong, and C.~C. Loy, ``Online deep clustering for
  unsupervised representation learning,'' in \emph{Proceedings of IEEE/CVF
  Conference on Computer Vision and Pattern Recognition (CVPR)}, 2020.

\bibitem{chaitanya2020contrastive}
K.~Chaitanya, E.~Erdil, N.~Karani, and E.~Konukoglu, ``Contrastive learning of
  global and local features for medical image segmentation with limited
  annotations,'' in \emph{Neural Information Processing Systems (NeurIPS)},
  2020.

\bibitem{Kornblith2019Do}
S.~{Kornblith}, J.~{Shlens}, and Q.~V. {Le}, ``Do better imagenet models
  transfer better?'' in \emph{2019 IEEE/CVF Conference on Computer Vision and
  Pattern Recognition (CVPR)}, 2019, pp. 2656--2666.

\bibitem{tajbakhsh2016convolutional}
N.~Tajbakhsh, J.~Y. Shin, S.~R. Gurudu, R.~T. Hurst, C.~B. Kendall, M.~B.
  Gotway, and J.~Liang, ``Convolutional neural networks for medical image
  analysis: Full training or fine tuning?'' \emph{IEEE transactions on medical
  imaging}, vol.~35, no.~5, pp. 1299--1312, 2016.

\bibitem{Neyshabur2020What}
B.~Neyshabur, H.~Sedghi, and C.~Zhang, ``What is being transferred in transfer
  learning?'' in \emph{Neural Information Processing Systems (NeurIPS)}, 2020.

\bibitem{yosinski2014transferable}
J.~Yosinski, J.~Clune, Y.~Bengio, and H.~Lipson, ``How transferable are
  features in deep neural networks?'' in \emph{Advances in neural information
  processing systems}, 2014, pp. 3320--3328.

\bibitem{raghu2019transfusion}
M.~Raghu, C.~Zhang, J.~Kleinberg, and S.~Bengio, ``Transfusion: Understanding
  transfer learning with applications to medical imaging,'' \emph{arXiv
  preprint arXiv:1902.07208}, 2019.

\bibitem{wang2019e2train}
Y.~Wang, Z.~Jiang, X.~Chen, P.~Xu, Y.~Zhao, Y.~Lin, and Z.~Wang, ``E2-train:
  Training state-of-the-art cnns with over 80

\bibitem{Amber2019large}
\BIBentryALTinterwordspacing
A.~L. Simpson, M.~Antonelli, S.~Bakas, M.~Bilello, K.~Farahani, B.~van
  Ginneken, A.~Kopp{-}Schneider, B.~A. Landman, G.~Litjens, B.~H. Menze,
  O.~Ronneberger, R.~M. Summers, P.~Bilic, P.~F. Christ, R.~K.~G. Do,
  M.~Gollub, J.~Golia{-}Pernicka, S.~Heckers, W.~R. Jarnagin, M.~McHugo,
  S.~Napel, E.~Vorontsov, L.~Maier{-}Hein, and M.~J. Cardoso, ``A large
  annotated medical image dataset for the development and evaluation of
  segmentation algorithms,'' \emph{CoRR}, vol. abs/1902.09063, 2019. [Online].
  Available: \url{http://arxiv.org/abs/1902.09063}
\BIBentrySTDinterwordspacing

\bibitem{hendrycks2019using}
D.~Hendrycks, M.~Mazeika, S.~Kadavath, and D.~Song, ``Using self-supervised
  learning can improve model robustness and uncertainty,'' in \emph{Advances in
  Neural Information Processing Systems}, 2019, pp. 15\,637--15\,648.

\bibitem{caron2019unsupervised}
M.~Caron, P.~Bojanowski, J.~Mairal, and A.~Joulin, ``Unsupervised pre-training
  of image features on non-curated data,'' in \emph{Proceedings of IEEE
  International Conference on Computer Vision}, 2019, pp. 2959--2968.

\bibitem{zhang2019aet}
L.~Zhang, G.-J. Qi, L.~Wang, and J.~Luo, ``Aet vs. aed: Unsupervised
  representation learning by auto-encoding transformations rather than data,''
  in \emph{Proceedings of the IEEE Conference on Computer Vision and Pattern
  Recognition}, 2019, pp. 2547--2555.

\bibitem{Newell2020How}
A.~Newell and J.~Deng, ``How useful is self-supervised pretraining for visual
  tasks?'' in \emph{Proceedings of the IEEE/CVF Conference on Computer Vision
  and Pattern Recognition (CVPR)}, June 2020.

\bibitem{Lowe2004Distinctive}
D.~G. Lowe, ``Distinctive image features from scale-invariant keypoints,''
  \emph{International Journal of Computer Vision}, vol.~60, pp. 91--110, 2004.

\bibitem{Ng2015Exploiting}
J.~Yue-Hei~Ng, F.~Yang, and L.~S. Davis, ``Exploiting local features from deep
  networks for image retrieval,'' in \emph{Proceedings of IEEE Conference on
  Computer Vision and Pattern Recognition (CVPR) Workshops}, 2015.

\bibitem{Gidaris2020Learning}
S.~Gidaris, A.~Bursuc, N.~Komodakis, P.~Perez, and M.~Cord, ``Learning
  representations by predicting bags of visual words,'' in \emph{Proceedings of
  IEEE/CVF Conference on Computer Vision and Pattern Recognition (CVPR)}, 2020.

\bibitem{Arandjelovic2016NetVLAD}
R.~Arandjelovic, P.~Gronat, A.~Torii, T.~Pajdla, and J.~Sivic, ``Netvlad: Cnn
  architecture for weakly supervised place recognition,'' in \emph{Proceedings
  of the IEEE Conference on Computer Vision and Pattern Recognition (CVPR)},
  June 2016.

\bibitem{gansbeke2020scan}
W.~V. Gansbeke, S.~Vandenhende, S.~Georgoulis, M.~Proesmans, and L.~V. Gool,
  ``Scan: Learning to classify images without labels,'' 2020.

\bibitem{caron2018deep}
M.~Caron, P.~Bojanowski, A.~Joulin, and M.~Douze, ``Deep clustering for
  unsupervised learning of visual features,'' in \emph{Proceedings of the
  European Conference on Computer Vision}, 2018, pp. 132--149.

\bibitem{ross2018exploiting}
T.~Ross, D.~Zimmerer, A.~Vemuri, F.~Isensee, M.~Wiesenfarth, S.~Bodenstedt,
  F.~Both, P.~Kessler, M.~Wagner, B.~M{\"u}ller \emph{et~al.}, ``Exploiting the
  potential of unlabeled endoscopic video data with self-supervised learning,''
  \emph{International journal of computer assisted radiology and surgery},
  vol.~13, no.~6, pp. 925--933, 2018.

\bibitem{Bai2019Self}
W.~Bai, C.~Chen, G.~Tarroni, J.~Duan, F.~Guitton, S.~E. Petersen, Y.~Guo, P.~M.
  Matthews, and D.~Rueckert, ``Self-supervised learning for cardiac mr image
  segmentation by anatomical position prediction,'' in \emph{Medical Image
  Computing and Computer Assisted Intervention -- MICCAI 2019}, D.~Shen,
  T.~Liu, T.~M. Peters, L.~H. Staib, C.~Essert, S.~Zhou, P.-T. Yap, and
  A.~Khan, Eds.\hskip 1em plus 0.5em minus 0.4em\relax Cham: Springer
  International Publishing, 2019, pp. 541--549.

\bibitem{haghighi2020learning}
F.~Haghighi, M.~R. Hosseinzadeh~Taher, Z.~Zhou, M.~B. Gotway, and J.~Liang,
  ``Learning semantics-enriched representation via self-discovery,
  self-classification, and self-restoration,'' in \emph{Medical Image Computing
  and Computer Assisted Intervention -- MICCAI 2020}.\hskip 1em plus 0.5em
  minus 0.4em\relax Cham: Springer International Publishing, 2020, pp.
  137--147.

\end{thebibliography}

\newpage

\appendices
\section{Supplemental materials}

\subsection{Target tasks and datasets}
\label{appendix_dataset}
We have evaluated TransVW in seven distinct target tasks, including classification and segmentation in CT, MRI, and X-ray modalities, detailed as follows.

\smallskip
\noindent\textbf{Lung nodule false positive reduction (\texttt{NCC}):}
LUNA16~\cite{setio2017validation} dataset provides 888 low-dose lung CT scans with a slice thickness of less than 2.5mm, divided into a training set (445 cases), a validation set (178 cases), and a test set (265 cases).
The dataset offers the annotations for a set of 5M candidate locations for the false positive reduction task, wherein true positives are labeled as ``1'' and false positives are labeled as ``0''. Area Under the Curve (AUC) score on classifying true positives and false positives is utilized as the evaluation metric.

\smallskip
\noindent\textbf{Lung nodule segmentation (\texttt{NCS}):}
We have evaluated TransVW for lung nodule segmentation using Lung Image Database Consortium image collection (LIDC-IDRI)~\cite{armato2011lung} dataset. This 
 dataset provides 1,018 thoracic CT scans with marked-up annotated lung nodules created by seven academic centers and eight medical imaging companies. The dataset is splitted into training (510), validation (100), and test (408) sets. 
We have re-sampled the 3D volumes to 1-1-1 spacing and then extracted a 64$\times$64$\times$32 crop around each nodule. These 3D crops are used for model training and evaluation. Intersection over Union (IoU) and  Dice coefficient scores are utilized to evaluate the lung nodule segmentation performance.

\smallskip
\noindent\textbf{Pulmonary embolism false positive reduction (\texttt{ECC}):} A database consisting of 121 computed tomography pulmonary angiography (CTPA) scans with a total of 326 emboli were collected in~\cite{tajbakhsh2015computer}, and divided at the patient-level into a training set with 434 true positive and 3,406 false positive PE candidates, and a test set with 253 true positive PE candidates and 2,162 false positive PE candidates.
The dataset is pre-processed as suggested in~\cite{zhou2019models}. The classification of true positives and false positives is evaluated by candidate-level AUC.

\smallskip
\noindent\textbf{Liver segmentation (\texttt{LCS}):}
We have utilized the dataset provided by MICCAI 2017 LiTS Challenge for evaluating TransVW on liver segmentation task. This dataset
consists of 130 CT scans, with the segmentation annotations for liver and lesion.  In our experiments, we split dataset into training (100 patients), validation (15 patients), and test (15 patients) sets, and consider liver as positive class and others as negative class. Segmentation performance is evaluated by Intersection over Union (IoU) and Dice coefficient scores.

\smallskip
\noindent\textbf{Brain tumor segmentation (\texttt{BMS}):} 
We have examined TransVW for brain tumor segmentation task on MRI Flair images provided by Brain Tumor segmentation (BraTS) 2018 dataset~\cite{bakas2018identifying}. 
This dataset provides 285 patients (210 HGG and 75 LGG), each with four different MR volumes including native
T1-weighted (T1), post-contrast T1-weighted (T1Gd), T2-weighted (T2), and T2 fluid attenuated inversion recovery (FLAIR). Segmentation annotations are provided for 
background (label 0), GD-enhancing tumor (label 4), the peritumoral
edema (label 2), and the necrotic and non-enhancing tumor core (label 1).
We split the data to 190 patients for training and 95 patients for testing. We consider background as negatives class and tumor sub-regions as positive class, and evaluate segmentation performance using Intersection over Union (IoU) and Dice coefficient scores.

\smallskip
\noindent\textbf{Thorax diseases classification (\texttt{DXC}):}
ChestX-ray14~\cite{wang2017chestx} is a hospital-scale chest X-ray dataset, which consists of 112K frontal-view X-ray images taken from 30K patients where 51K images have at least one of the 14 thorax diseases. ChestX-ray14 provides an  patient-wise split for training (86K images) and test (25K images) sets with 14 disease labels (each image can have multi-labels). 
We report the mean AUC score over 14 diseases for the multi-label chest X-ray classification task.

\smallskip
\noindent\textbf{Pneumothorax Segmentation (\texttt{PXS}):}
The Society for Imaging Informatics in Medicine (SIIM) and American College of Radiology provided the SIIM-ACR Pneumothorax Segmentation dataset~\cite{PNEchallenge}, consisting of 10K chest X-ray images and the segmentation masks for Pneumothorax disease. We divide the dataset into training (8K), validation (1K), and testing (2K), and evaluate the segmentation performance using Dice coefficient score.

\subsection{Impact of number of visual words classes}
\label{appendix_abl_num_vws}

\begin{figure*}[t]
\centerline{\includegraphics[width=\textwidth]{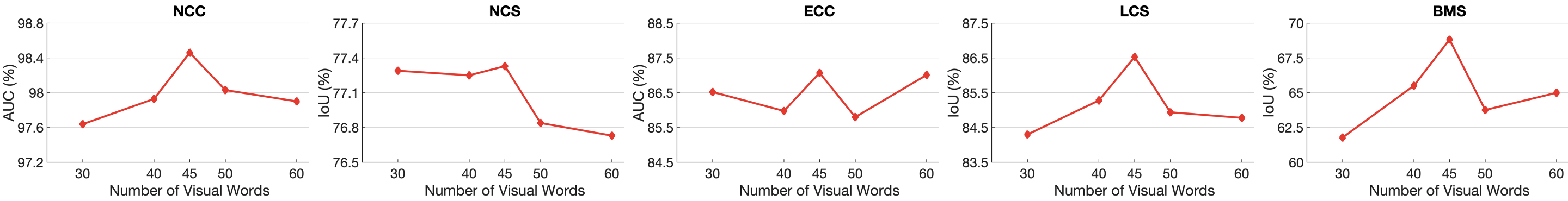}}
\caption{We conduct ablation study on the impact of number of visual words classes on the target task performance on five 3D target tasks. We report the average performance over ten runs for each model on each task. The best performance  achieved with $C=45$ in all applications.   }
\label{fig:ablation_nb_vws}
\end{figure*}

To explore the impact of number of visual words classes ($C$) on the performance of target tasks, we have
conducted extensive ablation studies on the number of classes. \figurename~\ref{fig:ablation_nb_vws} shows the performance of TransVW on  all five 3D target tasks under different settings. We report the average performance over ten runs for each model on each application. The best performance achieved at $C=45$ in all applications.  
We suggest that for achieving the best transfer learning performance, it is necessary to strike a balance between diversity and overlap of the visual words.

\subsection{Visual Words Visualization}
\label{appendix_vws}
\begin{figure*}[t]
\centerline{\includegraphics[width=\textwidth]{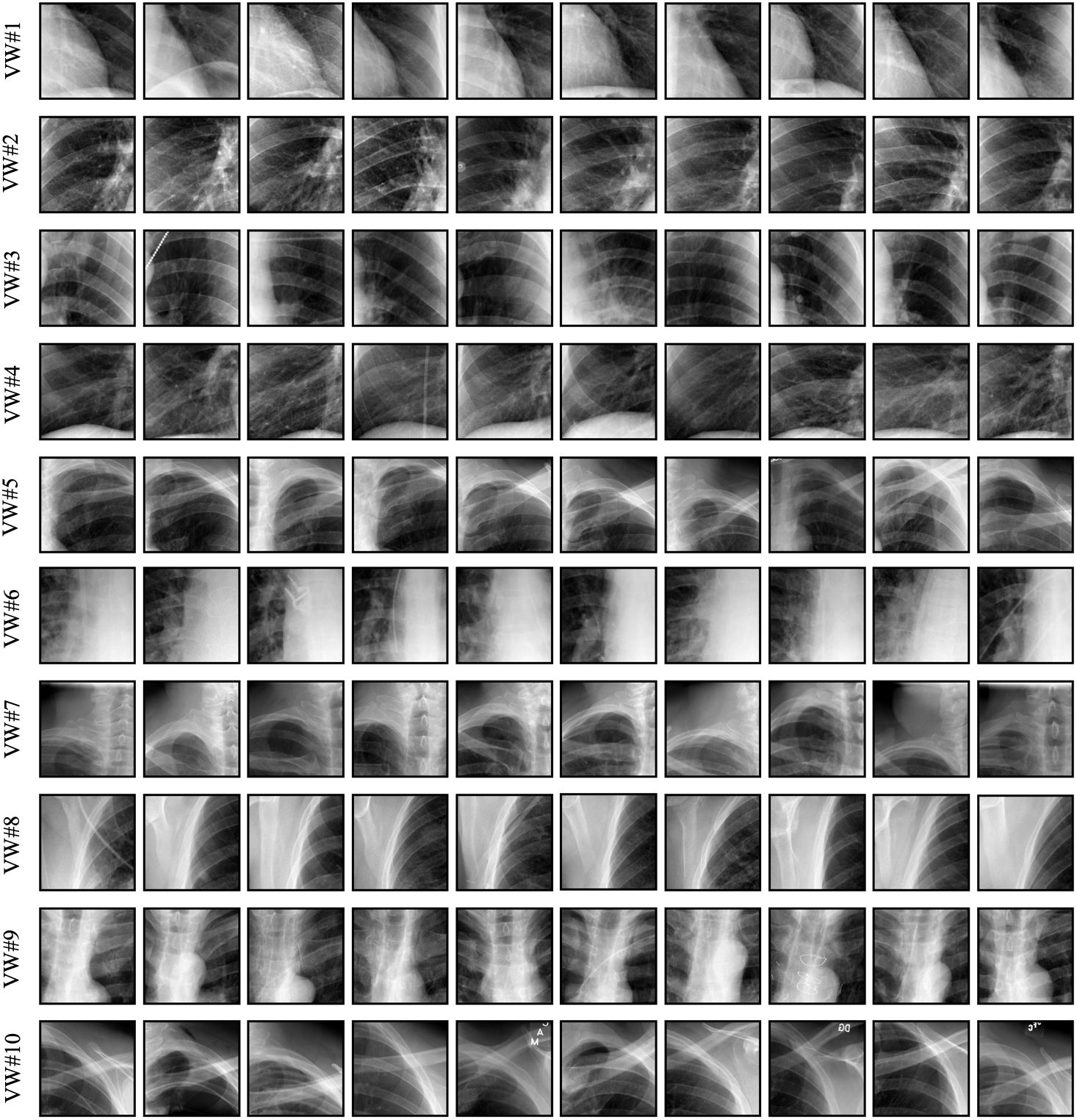}/}
\caption{Visualization of visual words in X-ray images.  Each row presents ten instances of a distinct visual word that are extracted from ten examples randomly selected from 1,000 nearest neighbors to a random reference image, to which we assign labels 1-10.}
\label{fig:vws_1_10}
\end{figure*}

We devise a self-discovery scheme to automatically extract visual words directly from unlabeled medical images images, resulting in a well-balanced and diversified dataset associated with semantically meaningful labels. As an example, in \figurename~\ref{fig:vws_1_10}, we present instances of ten visual words.  As seen, each visual word covers a specific anatomical pattern which is recurrent across all images, to which we assign labels 1-10.

\subsection{Visualizing the self-discovery process}
\label{appendix_self_discovery}
\begin{figure*}[t]

\begin{center}
 \includegraphics[width=0.9\textwidth]{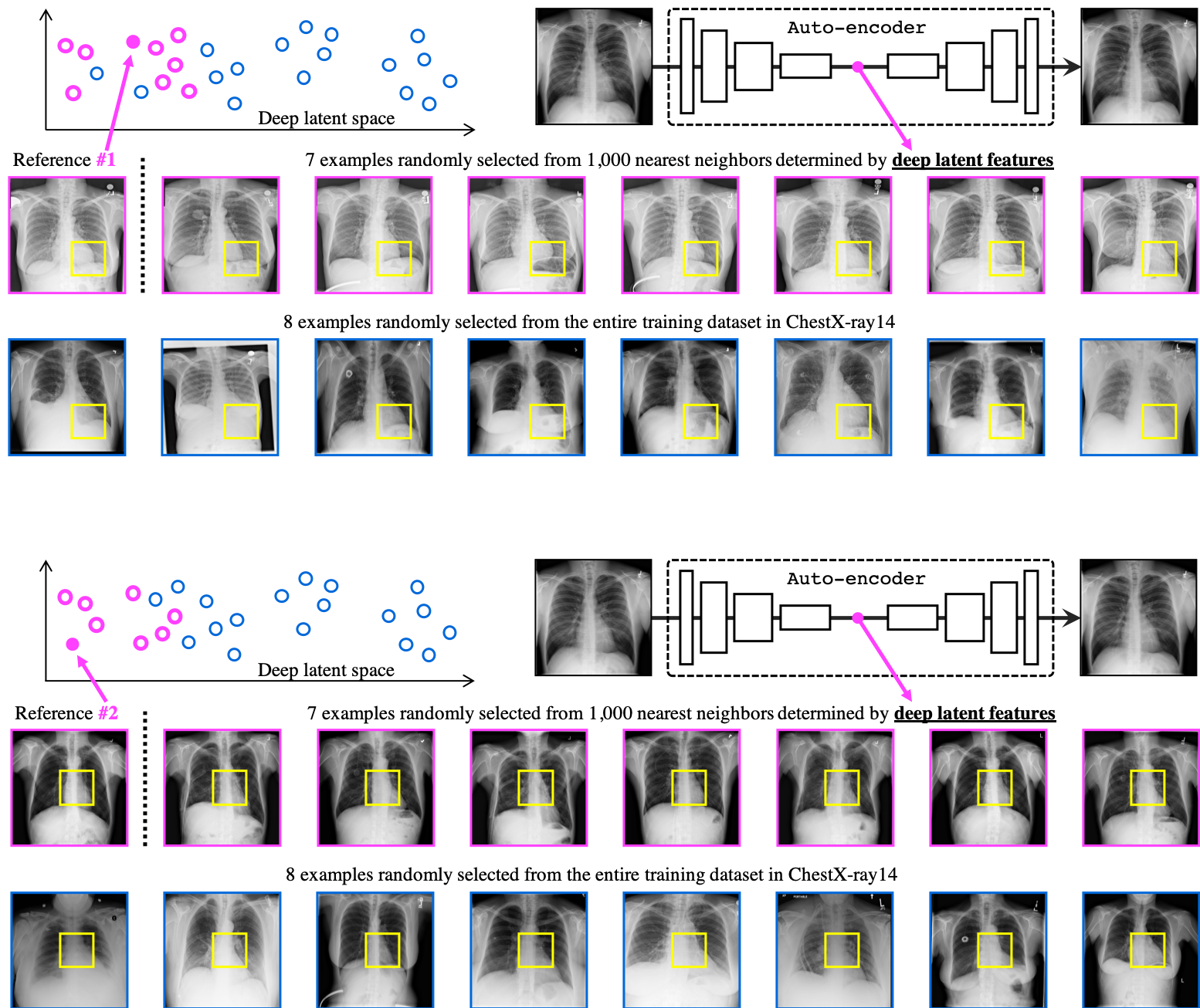}
\end{center}
  \caption{
  Without loss of generalization, and for simplicity and clarity, we present our idea with X-ray images.  Our self-discovery process seeks to automatically discover similar visual words across patients, as illustrated in the yellow boxes within the patients framed
  in pink. Patches extracted at the same coordinate across patients may be very different
(the yellow boxes within the patients framed in blue). We overcome this issue by first
computing similarity at the patient level using the deep latent features from a feature extractor network pre-trained with an unsupervised task (\eg an autoencoder) and then mining the top nearest neighbors (framed in pink) of a random reference
patient. Extracting visual word instances from these similar patients strikes a balance
between consistency and diversity in pattern appearance for each visual word.
  }
\label{fig:appendix_discover_patients}
\end{figure*}

To build a more comprehensive understanding of the proposed self-discovery scheme, we randomly anchor two patients as references and visualize the self-discovery process in \figurename~\ref{fig:appendix_discover_patients}.

\subsection{Incorporating TransVW with nnU-Net framework}
\label{appendix_discussion_nnunet}
The nnU-Net framework~\cite{isensee2020automated} has shown state-of-the-art performance in various segmentation  tasks in medical imaging; specifically, it won first place in the 2018 Decathlon challenge~\cite{Amber2019large}. 
To further demonstrate the transferability and effectiveness of our pre-trained model, we have adopted TransVW to the nnU-Net framework and evaluated  it  on  liver  tumor  segmentation  task from the Medical Segmentation Decathlon challenge. To do so, we have pre-trained TransVW on liver architecture from nn-UNet and then fine-tuned it with the provided training data from the challenge without using external data.  Based on our experimental results in this task, we have observed the following:

\noindent 1) TransVW yields a 1\% boost to the Dice score upon fine-tuning on the test set compared with the nnU-Net model trained from scratch. At the time of manuscript submission, fine-tuning TransVW beats nnU-Net trained from scratch according to the official scores on the live leaderboard\footnote{\label{foot:leaderboard}Online leaderboard of Decathlon challenge: \href{https://decathlon-10.grand-challenge.org/evaluation/challenge/leaderboard/}{https://decathlon-10.grand-challenge.org/evaluation/challenge/leaderboard/}}, 0.77 vs. 0.76. 

\noindent 2) TransVW obtains improvements over training from scratch and Models Genesis by 2.5\% and 1\%,  respectively, to the Dice scores. Since the labels for the test images are not publicly available, the evaluation was conducted by five-fold cross-validation on the training dataset as in~\cite{isensee2020automated}. To conduct fair comparisons, all methods benefit from the same data augmentation and use the same network architecture (nnU-Net architecture for liver), while we endeavor to optimize each model with the best-performing hyper-parameters. Following this evaluation protocol, learning from scratch, Models Genesis, and TransVW achieve a Dice score of 63.52\%$\pm0.28$\%, 65.03\%$\pm0.70$\%, and 66.06\%$\pm0.98$\%, respectively. 

\noindent 3) TransVW has proven to be a strong contender for first place in the live leaderboard of the challenge. At the time of manuscript submission, the Dice score achieved by TransVW as well as the top-ranking model (\textit{peggyko}) was 0.77; the leaderboard only reports up to two significant digits.

Overall, these results reinforce our main insight that TransVW can serve as a primary source of transfer learning for 3D medical imaging applications to boost performance, accelerate training, and reduce annotation costs, representing its clinical significance.

\end{document}